\documentclass[10pt,twocolumn,letterpaper]{article}

\usepackage[pagenumbers]{wacv} 

\usepackage{tabularray}
\usepackage{adjustbox}
\usepackage{dsfont}
\usepackage{float}

\definecolor{wacvblue}{rgb}{0.21,0.49,0.74}
\usepackage[pagebackref,breaklinks,colorlinks,allcolors=wacvblue]{hyperref}

\def\wacvPaperID{1305} 
\def\confName{WACV}
\def\confYear{2026}

\title{Performance of Conformal Prediction in Capturing Aleatoric Uncertainty}

\author{
Misgina Tsighe Hagos$^{1}$ \quad Claes Lundström$^{1,2}$\\
$^{1}$Linköping University\\
$^{2}$Sectra\\
{\tt\small \{misgina.tsighe.hagos\}@liu.se}
}

\begin{document}
\maketitle
\begin{abstract}

Conformal prediction is a model-agnostic approach to generating prediction sets that cover the true class with a high probability. Although its prediction set size is expected to capture aleatoric uncertainty, there is a lack of evidence regarding its effectiveness. The literature presents that prediction set size can upper-bound aleatoric uncertainty or that prediction sets are larger for \emph{difficult} instances and smaller for \emph{easy} ones, but a validation of this attribute of conformal predictors is missing. This work investigates how effectively conformal predictors quantify aleatoric uncertainty, specifically the inherent ambiguity in datasets caused by overlapping classes. We perform this by measuring the correlation between prediction set sizes and the number of distinct labels assigned by human annotators per instance. We further assess the similarity between prediction sets and human-provided annotations. We use three conformal prediction approaches to generate prediction sets for eight deep learning models trained on four datasets. The datasets contain annotations from multiple human annotators (ranging from five to fifty participants) per instance, enabling the identification of class overlap. We show that the vast majority of the conformal prediction outputs show a very weak to weak correlation with human annotations, with only a few showing moderate correlation. These findings underscore the necessity of critically reassessing the prediction sets generated using conformal predictors. While they can provide a higher coverage of the true classes, their capability in capturing aleatoric uncertainty and generating sets that align with human annotations remains limited.

\end{abstract}
    
\section{Introduction}
\label{sec:intro}

Quantifying model uncertainty is critical for improving model transparency and can enhance the performance of downstream tasks such as active learning and out-of-distribution (OOD) detection \cite{mucsanyi2023trustworthy,mucsanyi2024benchmarking}. It is usually categorized into aleatoric uncertainty (inherent to the data and irreducible) and epistemic uncertainty (related to model choice and reducible) to satisfy each task's requirements \cite{senge2014reliable,kendall2017uncertainties,hullermeier2021aleatoric,korchagin2025improving,cocheteux2025uncertainty}. For example, in active learning, we would want to deprioritize samples with high aleatoric and low epistemic uncertainty \cite{mukhoti2023deep}. Similarly, in OOD detection, we want to avoid confounding previously seen (in-distribution) high aleatoric uncertainty samples with high epistemic uncertainty samples.

Conformal prediction (CP) is a framework for uncertainty quantification that generates prediction sets that are statistically guaranteed to include the true class at a chosen confidence level \cite{vovk2005algorithmic,romano2020classification}. This is as opposed to the conventional models that output maximum-likelihood predictions $y \in \mathcal{Y}$. CP is particularly helpful where class distributions overlap and we need to identify all the likely classes and rule out the rest \cite{angelopoulosuncertainty}.

For a model trained on a dataset $(X_i, Y_i) \in \mathcal{X} \times \mathcal{Y}$ for $i = 1, \dots, n$, CP constructs a prediction set $\mathcal{C}(X_{\text{test}}) \subseteq \mathcal{Y}$ for a test input $(X_{\text{test}}, Y_{\text{test}})$, with the finite-sample guarantee, $\mathbb{P}(Y_{\text{test}} \in \mathcal{C}(X_{\text{test}})) \geq 1 - \alpha$, where $\alpha \in (0, 1)$ is a user-specified target error level.




CP captures aleatoric uncertainty by calibrating prediction sets whose size adapts to the ambiguity in the data, regardless of the underlying model architecture \cite{sadinle2019least,clark2024conformal,cabezas2025epistemic}. This ability stems from using a conformity score, often derived from model outputs such as softmax probabilities, whose entropy quantifies aleatoric uncertainty on a per-instance basis \cite{mukhoti2023deep}. Softmax outputs can be overconfident and poorly aligned with true likelihoods, which can degrade the quality of CP sets \cite{huang2023conformal}. Consequently, assessing their calibration, typically via metrics such as the Expected Calibration Error (ECE), is essential before using them for CP. Given class probabilities that are log-concave and that achieve near-perfect calibration, we can assume that the prediction set size, $|\mathcal{C}(X)|$, and an existing aleatoric uncertainty quantification method, softmax entropy \cite{mukhoti2023deep}, are positively correlated. We show evidence of this in the later parts of the paper. \citet{correia2024information} also provide theoretical evidence that the prediction set size upper bounds the conditional entropy $H(X | Y)$, which captures the true intrinsic uncertainty in the distribution of labels $Y$ and inputs $X$. However, there is no clear empirical evidence of how good conformal predictors are at capturing aleatoric uncertainty, or if they do at all. 


Beyond theoretical interest, understanding what prediction set sizes represent is essential for their effective use in downstream tasks, particularly in decision-support systems that involve human interaction. User studies have shown that CP can improve human-AI complementarity in automated decision-support systems \cite{straitouri2023improving,straitouri2024designing,cresswell2024conformal}. In such experiments, the size of the prediction sets is assumed to determine the difficulty of a classification task for human participants. This paper evaluates this assumption empirically. 



In this work, we specifically focus on aleatoric uncertainty stemming from class overlap, i.e., cases where semantically distinct classes exhibit similar visual or structural characteristics in the input space, rendering accurate classification inherently uncertain. This form of uncertainty is often revealed through human annotation studies, where multiple annotators are tasked with labeling the same instance. When annotators consistently assign multiple distinct labels to a single instance, either individually or across annotators, this pattern suggests the presence of class overlap. It indicates that the resulting uncertainty is not due to labeling error, but rather reflects intrinsic ambiguity in the data itself \cite{marchal2022establishing,uma2022scaling}.

To systematically capture this form of instance-level ambiguity, several datasets have been re-annotated---either in whole or in part---with labels from multiple annotators to better reflect human perception and uncertainty \cite{peterson2019human, qi2020mlrsnet, beyer2020we}. Ambiguous instances are typically characterised by multiple, usually conflicting, labels assigned by different annotators, while non-ambiguous instances often tend to receive similar and consistent annotations \cite{peterson2019human, kirchhof2023probabilistic}. For model training purposes, single labels are often derived via majority voting or by retaining the original labels, whereas the full distribution of annotator labels is preserved in the test set. This allows for a more nuanced evaluation of uncertainty quantification frameworks such as CP, particularly in capturing instance-level ambiguity.


In this paper, we focus on datasets that contain multiple human annotations per instance. This enables us to identify and quantify class overlap as a source of aleatoric uncertainty. In addition to class overlap, we also consider the distribution of human annotations whenever such information is available. We then investigate whether the human-perceived ambiguity correlates with the uncertainty captured by conformal predictors, as reflected in the size of their prediction sets. While annotator disagreement has also been used as a proxy for aleatoric uncertainty \cite{tran2022plex,kirchhof2023probabilistic,mucsanyi2024benchmarking}, we focus on class overlap since we are interested in assessing the prediction sets. We also evaluate the performance of these prediction sets and analyze how their size influences both coverage and the strength of the correlation with class overlap. Additionally, we employ similarity metrics to assess if the prediction sets correctly identify labels collected from the multiple annotators. We define an instance as exhibiting class overlap if it contains two or more distinct class labels from an individual or multiple annotators. The degree of class overlap is quantified by the number of distinct labels assigned to the instance.

The primary contributions of our paper are as follows,

\begin{itemize}
    \item We evaluate the performance of three CP approaches at capturing aleatoric uncertainty caused by an inherent class overlap in four datasets.
    \item We show that the employed conformal predictors frequently construct prediction sets that correlate weakly with the datasets' inherent class overlap. 
    \item We provide empirical evidence that a growing prediction set size does not necessarily result in improved coverage of the true classes, especially in datasets with a large number of categories.
    \item In addition, we also find that most of the prediction sets substantially differ from human annotations.
\end{itemize}

\section{Related work}
\label{sec:rel_work}

Despite achieving high predictive accuracy, deep learning models often produce overconfident softmax probabilities, making them unreliable for uncertainty estimation \cite{guo2017calibration}. In response, a range of model uncertainty quantification methods have been proposed, including Monte Carlo dropout \cite{gal2016dropout}, test-time augmentation \cite{ayhan2018test}, deep ensembles \cite{lakshminarayanan2017simple}, and evidential deep learning \cite{sensoy2018evidential}. While effective in some settings, these methods often suffer from architectural constraints and substantial computational overhead \cite{cresswell2024conformal}. As an alternative, CP offers a model-agnostic approach that mirrors human behavior under uncertainty, providing sets of plausible answers rather than a single maximum-likelihood prediction \cite{vovk2005algorithmic,romano2020classification}. These prediction sets adapt in size to reflect the model's confidence, offering a direct and interpretable measure of uncertainty \cite{sadinle2019least,clark2024conformal,cabezas2025epistemic}.

Uncertainty quantification methods are commonly evaluated through proxy tasks such as active learning and OOD detection \cite{mukhoti2023deep}. While useful, these indirect assessments often rely on task-specific assumptions that may not align with the original objective of uncertainty estimation. An alternative is directly evaluating uncertainty estimates on the primary task, e.g., classification, by comparing them to human annotator disagreement \cite{schrufer2024you}. This provides a task-relevant benchmark that more faithfully reflects the intrinsic ambiguity present in the data. In contrast to active learning, which evaluates uncertainty based on its utility for informative sample selection rather than its fidelity to actual data ambiguity, or OOD detection, which relies on model confidence to flag inputs unlike the training data rather than genuine data ambiguity, human annotators' based evaluation offers a principled target for evaluating uncertainty grounded in the data's inherent complexity.

While ambiguity in human annotations has been proposed as a benchmark for model uncertainty, conformal predictors are typically assessed solely based on their coverage guarantees, i.e., whether the true class falls within the prediction set \cite{angelopoulosuncertainty,clark2024conformal}, without regard to whether their prediction sets align with regions of genuine ambiguity in the data. Although coverage is central to the CP framework, it does not capture the informativeness of the prediction sets themselves. 


\section{Conformal prediction methods}
\label{sec:cp_methods}

In multi-class classification, we consider a predictive model $\hat{f} : \mathcal{X} \rightarrow \mathbb{R}^K$ that outputs a confidence vector $\hat{f}(X) \in \mathbb{R}^K$ for each input $X \in \mathcal{X}$, where $\hat{f}(X)y$ denotes the predicted probability assigned to class $y \in {1, \dots, K}$. For a test input $x_{\text{test}}$, we construct a prediction set $\mathcal{C}(x_{\text{test}}) \subseteq {1, \dots, K}$ that contains all classes satisfying:

\begin{equation}
\mathcal{C}(x_{\text{test}}) = \left\{ y \in {1, \dots, K} : \hat{f}(x_{\text{test}})_y \geq 1 - \hat{q} \right\},
\end{equation}

\noindent where $\hat{q}$ is a threshold computed from a calibration set $(X_{\text{cal}}, Y_{\text{cal}})$ under the assumption that the calibration and test sets are exchangeable. 

We use three CP approaches to generate prediction sets: the least ambiguous set-valued classifier (LAC) \cite{sadinle2019least}, adaptive prediction sets (APS) \cite{romano2020classification}, and regularized adaptive prediction sets (RAPS) \cite{angelopoulosuncertainty}. They are introduced next.

\paragraph{Least Ambiguous Set-Valued Classifier}

The Least Ambiguous Classifier (LAC) defines a score function,

\begin{equation}
    s(x,y) = 1 - \hat{f}(x)_{y},
\end{equation}

\noindent where $\hat{f}(x)_y$ is the softmax probability assigned to the true class $y$. For a calibration set of size $n$ and target error rate $\alpha$, the $(1-\alpha)$ quantile is computed applying a finite-sample correction:

\begin{equation}
    \hat{q} = \text{quantile} \left( \{s_1, \dots, s_n\}; \frac{\lceil(1 - \alpha)(n + 1)\rceil}{n} \right),
\end{equation}

\noindent The prediction set is then constructed:

\begin{equation}
    \mathcal{C}(x_{\text{test}}) = \left\{ y : \hat{f}(x_{\text{test}})_y \geq 1 - \hat{q} \right\}
\end{equation}

Because it only considers the true class probability during calibration, LAC can yield empty prediction sets for high-uncertainty inputs.

\paragraph{Adaptive Prediction Sets}

Adaptive Prediction Sets (APS) sort the softmax outputs in descending order to obtain $\pi(x)$. The conformity score is the cumulative probability up to the true class index:

\begin{equation}
    s(x,y) = \sum_{j=1}^{k} \hat{f}(x)_{\pi_j(x)},
\end{equation}

\noindent where $\pi_k(x) = y$ and $k$ is the index of true class. Using the same quantile computation procedure as in LAC, the prediction set is given by:

\begin{equation}
\begin{split}
    \mathcal{C}(x_{\text{test}}) = \left\{ \pi_1, \dots, \pi_k \right\}, \quad \\
    k = \inf \left\{ k : \sum_{j=1}^{k} \hat{f}(x_{\text{test}})_{\pi_j} \geq \hat{q} \right\}.
\end{split}
\end{equation}

By summing over multiple class probabilities, APS avoids the empty prediction set issue observed in LAC.

\paragraph{Regularized Adaptive Prediction Sets}


RAPS augments APS with a regularization term that encourages prediction sets to reflect model confidence, while maintaining coverage. The conformity score is defined as:

\begin{equation}
    s(x,y) = \sum_{j=1}^{k} \hat{f}(x)_{\pi_j} + \lambda(k - k_{\text{reg}}),
\end{equation}

\noindent where $\pi_k = y$, and $k_{\text{reg}}$ is a target prediction set size. Following \citet{angelopoulosuncertainty}, we tune the regularization parameter $\lambda$ on a held-out 20\% subset of the calibration set, selecting from the grid $\{0.001, 0.01, 0.1, 0.2, 0.5\}$. This procedure yielded $\lambda = 0.1$ for all datasets, and we set the regularization parameter $k_{\text{reg}} = 1$.

After computing the quantile $\hat{q}$ as in LAC, the prediction set is constructed as:

\begin{equation}
\begin{split}
    \mathcal{C}(x_{\text{test}}) = \left\{ \pi_1, \dots, \pi_k \right\}, \quad \\ 
    k = \inf \left\{ k : \sum_{j=1}^{k} \hat{f}(x_{\text{test}})_{\pi_j} + \lambda(k - k_{\text{reg}}) \geq \hat{q} \right\}.
\end{split}
\end{equation}

\section{Experiments}
\label{sec:experiments}


This section presents a description of the datasets employed in this paper, the trained models, and the evaluation methods.

\subsection{Datasets}
\label{section:datasets}

\begin{figure}[htbp]
    \centering
    \adjustbox{trim=0pt 0.5pt 0pt 0pt, clip}{\includegraphics[trim=0 20 0 0, clip, width=0.47\columnwidth]{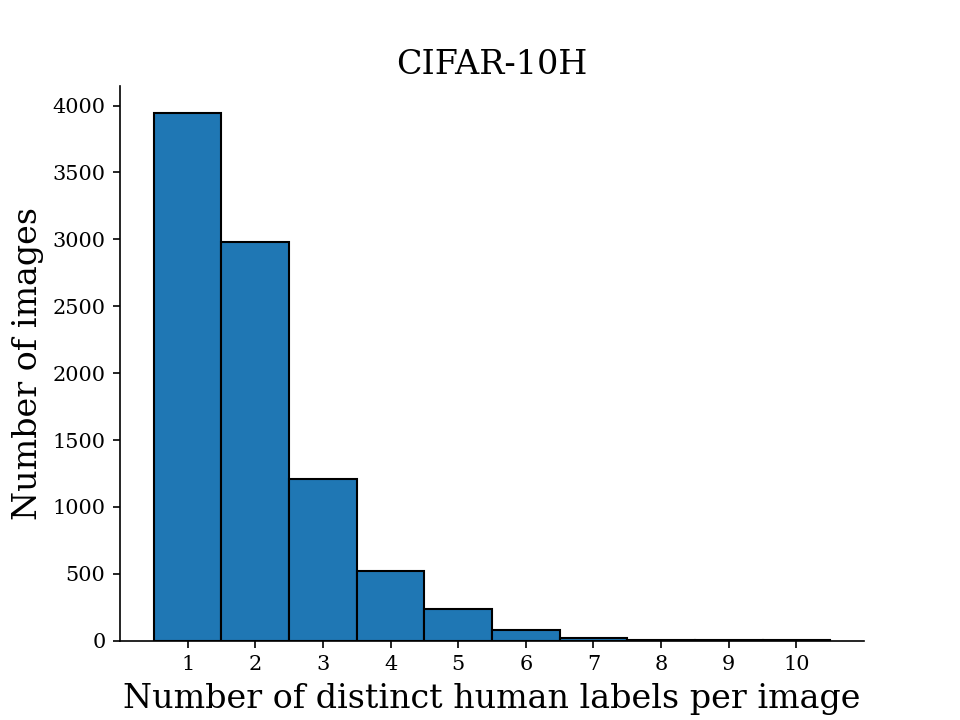}}
    \adjustbox{trim=5pt 5pt 0pt 0pt, clip}{\includegraphics[width=0.47\columnwidth]{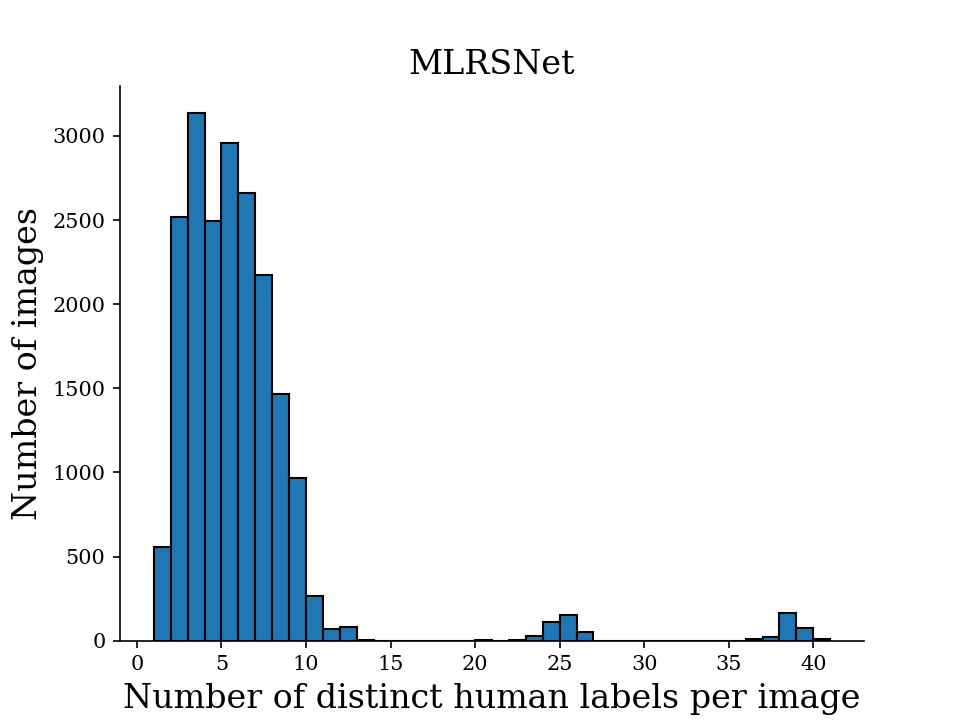}}
    \includegraphics[width=0.47\columnwidth ]{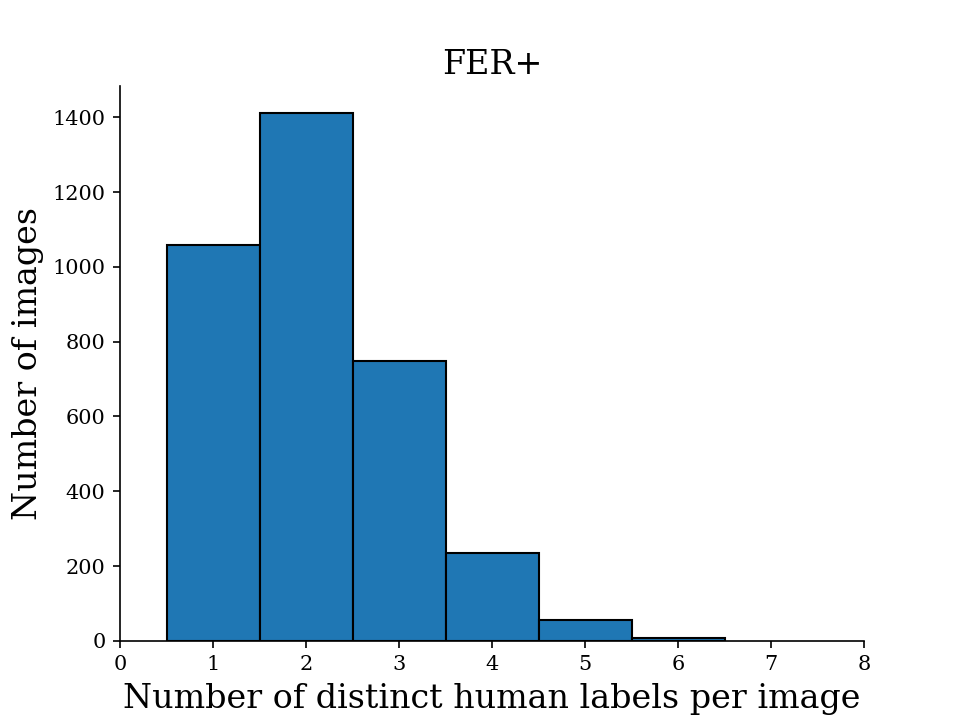}
    \adjustbox{trim=5pt 0pt 0pt 0pt, clip}{\includegraphics[width=0.47\columnwidth]{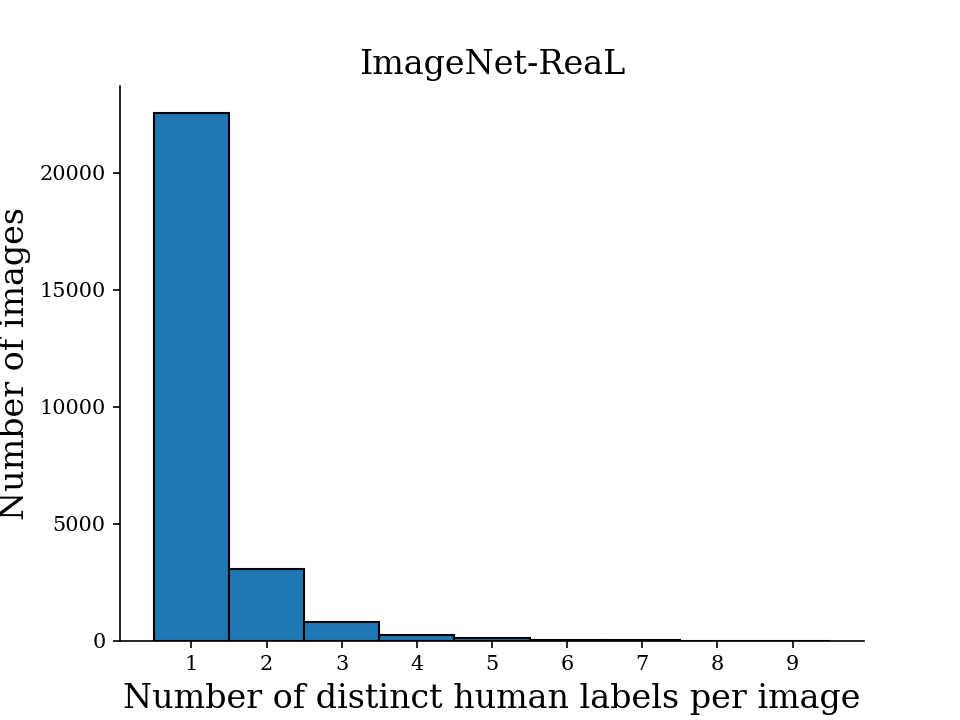}}
    \caption{Distinct human labels distribution in the CIFAR-10H, MLRSNet, FER+, and ImageNet-Real datasets. The X-axis shows the number of distinct human labels per image, and the Y-axis shows the number of images.}
    \label{figure:human_labels_distribution}
\end{figure}

The datasets used in our experiments are introduced next, with their corresponding count of distinct human labels distribution shown in Figure~\ref{figure:human_labels_distribution}. The average number of human annotators per instance is also reported when available in the datasets. For all the datasets, the calibration sets are selected from the same distribution as the test set.

\paragraph{CIFAR-10H} \cite{peterson2019human} is a collection of 511,400 human categorizations for the 10,000 images in the test subset of the CIFAR-10 dataset. We use 1,000 uniformly sampled instances from the original CIFAR-10 test set as a calibration set, with the remainder used for evaluation. An average of 50.09 annotators labelled each image in the test set, and class overlap exists in 56.2\% of these images. 

\paragraph{MLRSNet} \cite{qi2020mlrsnet} contains 46 categories. It contains 109,161 images, and labels are collected from 50 human annotators. We use 9,200 images, selected uniformly across all categories, as a calibration set, and 20,000 images as a test set. There is class overlap in 97.2\% of the test images.

\paragraph{FER+} \cite{BarsoumICMI2016} is a version of the facial expression recognition (FER) dataset \cite{dumitru2013challenges} that contains facial images and their corresponding single labels from seven categories. FER+ incorporates annotations from multiple human participants for each facial image. The dataset includes a training set of 28,709 images, and a public and private test set, each containing 3,589 images. We use the public and private test sets as calibration and test sets, respectively. On average, 9.35 annotators labelled the test images. Class overlap exists in 69.8\% of the test set images. 

\paragraph{ImageNet-ReaL} \cite{beyer2020we}  is constructed by re-annotating the ImageNet validation set \cite{russakovsky2015imagenet}, with each instance labeled by five human annotators. Each annotator is provided with eight candidate labels and asked whether each label is present in an image, setting the maximum number of labels for an image to eight. 15.8\% of the images received overlapping classes. After excluding images with missing labels, we selected 20,000 images from the validation set as the calibration set and 26,000 as the test set, ensuring a uniform distribution across all categories.

Per instance human annotation distribution is available for the CIFAR-10H and FER+ datasets. The rest of the datasets, i.e., MLRSNet and ImageNet-ReaL, only contain the distinct human-provided labels.

For each dataset, we searched for the best target error value $\alpha\in\{0.05, 0.1, 0.15, 0.2\}$ that minimizes the ratio of the CP mean set size to coverage. This ensures high coverage while keeping the prediction sets concise and informative. Apart from the ImageNet-ReaL dataset, which resulted in an $\alpha=0.10$, the rest of the datasets led to $\alpha=0.05$. In addition, the optimal calibration set size and thus the calibration/test split is determined by progressively enlarging the calibration set to improve coverage until it saturates.



\subsection{Models}

We used eight deep learning architecture: ResNet18, ResNet34, ResNet50, VGG-16, VGG-19, DenseNet121, DenseNet161, and MobileNet-v2.\footnote{\url{https://pytorch.org/vision/main/models.html}} All models were trained from scratch using an Adam optimizer with a decaying learning rate of 1e-4 and a categorical cross-entropy loss.


\subsection{Evaluation}

We use Spearman's rank correlation coefficient ($r_s$) to assess the correlation between the prediction set size and the class overlap or the number of distinct human labels. In addition, we compute Spearman's rank correlation between the prediction set size and human annotation distribution, using label entropy, in cases where label distribution information is available, i.e., for the CIFAR-10H and FER+ datasets. We assess the similarity between the prediction sets and human annotations using the similarity metrics precision, recall, subset-accuracy, and Hamming loss. Instances with empty prediction sets are excluded from the correlation and similarity analysis. 

We evaluate the performance of the conformal predictors using coverage, size-stratified coverage (SSC) \cite{angelopoulos2021gentle}, and mean prediction set size ($\bar{w}$). Coverage measures the ratio of the presence of the true class, $y$, in a prediction set, $\mathcal{C}(x)$, as follows,

\begin{equation}
    \text{Coverage} = \frac{1}{N} \sum_{i=1}^{N} \mathds{1} \left\{ y_i \in \mathcal{C}(x_i) \right\}
\end{equation}

SSC first discretizes the prediction set into $g$ bins based on set size and then computes the worst case coverage among all bins as follows,

\begin{equation} 
    \text{SSC} = \min_{g \in \{1, \dots, G\}} \frac{1}{|I_g|} \sum_{i \in I_g} \mathds{1} \left\{ y_i \in \mathcal{C}(x_i) \right\}
\end{equation}

\noindent where $I_g$ is a group of sets that belong to the $g^{th}$ size group, and $G$ is the number of distinct size groups. Mean set size assesses the capability of conformal predictors in generating concise sets. It is computed as, $\bar{w} = \frac{1}{N} \sum_{i=1}^{N} |\mathcal{C}(x_i)|$.


We compute Expected Calibration Error (ECE) \cite{naeini2015obtaining,guo2017calibration} to measure model calibration. ECE assesses the reliability of softmax outputs employed in constructing prediction sets. Lower ECE values indicate better calibration, with ECE=0 indicating perfect calibration and ECE=1 indicating maximum miscalibration. ECE first partitions predictions into M equally spaced bins, $B$, and takes a weighted average of the difference between accuracy and confidence. It is computed as follows for n instances.

\begin{equation}
\text{ECE} = \sum_{m=1}^{M} \frac{|B_m|}{n} \left| \text{acc}(B_m) - \text{conf}(B_m) \right|
\end{equation}
\section{Results}
\label{sec:results}

This section contains the results of our correlation and similarity analysis and the performance of the conformal prediction methods. Table~\ref{table:spearman_cp_against_human_labels} and Table~\ref{table:spearman_cp_against_softmax_entropy} present the correlation analysis results. Tables~\ref{table:coverage_only_performance_cifar10h}-\ref{table:avg_performance_w} contain the performance of the conformal predictors.

\subsection{Correlation analysis}
The Spearman's $r_s$ between prediction set sizes and class overlap is presented in Table~\ref{table:spearman_cp_against_human_labels}, and the Spearman's $r_s$ between the prediction set sizes and softmax entropy is shown in Table~\ref{table:spearman_cp_against_softmax_entropy}. We observe that Spearman's $r_s$ between prediction set size and class overlap is usually weak to very weak, with only a few showing a moderate correlation, indicating limited alignment of the prediction sets with human-perceived ambiguity. In contrast, as we expected, prediction set size usually exhibits a strong to very strong Spearman's $r_s$ with softmax entropy. 

An increased prevalence of larger prediction sets is associated with an improved Spearman's $r_s$ with class overlap, suggesting that prediction set size positively contributes to capturing instance-level ambiguity (See Figure~\ref{figure:spearman_correlation_against_prediction_set_sizes} in Supplementary material).

Although restricted to the CIFAR-10H and FER+ datasets, we also examine correlations using the label entropy in human annotations, as these are the only datasets that provide annotation distribution information. The Spearman's $r_s$ between prediction set size and human annotation entropy is reported in Table~\ref{table:spearman_cp_sets_and_human_annotation_entropy} in the Supplementary Material. Similar to the correlations against class overlap, the results are generally weak to very weak, with some cases showing moderate correlation.


\begin{table*}[h]
\centering
\caption{Spearman's rank correlation coefficient, $r_s$, $p < .001$, between conformal prediction set sizes and class overlap.}
\label{table:spearman_cp_against_human_labels}
\begin{tblr}{
  row{2} = {r},
  cell{1}{1} = {r=2}{},
  cell{1}{2} = {c=3}{c},
  cell{1}{5} = {c=3}{c},
  cell{1}{8} = {c=3}{c},
  cell{1}{11} = {c=3}{c},
  cell{3}{2} = {r},
  cell{3}{3} = {r},
  cell{3}{4} = {r},
  cell{3}{5} = {r},
  cell{3}{6} = {r},
  cell{3}{7} = {r},
  cell{3}{8} = {r},
  cell{3}{9} = {r},
  cell{3}{10} = {r},
  cell{3}{11} = {r},
  cell{3}{12} = {r},
  cell{3}{13} = {r},
  cell{4}{2} = {r},
  cell{4}{3} = {r},
  cell{4}{4} = {r},
  cell{4}{5} = {r},
  cell{4}{6} = {r},
  cell{4}{7} = {r},
  cell{4}{8} = {r},
  cell{4}{9} = {r},
  cell{4}{10} = {r},
  cell{4}{11} = {r},
  cell{4}{12} = {r},
  cell{4}{13} = {r},
  cell{5}{2} = {r},
  cell{5}{3} = {r},
  cell{5}{4} = {r},
  cell{5}{5} = {r},
  cell{5}{6} = {r},
  cell{5}{7} = {r},
  cell{5}{8} = {r},
  cell{5}{9} = {r},
  cell{5}{10} = {r},
  cell{5}{11} = {r},
  cell{5}{12} = {r},
  cell{5}{13} = {r},
  cell{6}{2} = {r},
  cell{6}{3} = {r},
  cell{6}{4} = {r},
  cell{6}{5} = {r},
  cell{6}{6} = {r},
  cell{6}{7} = {r},
  cell{6}{8} = {r},
  cell{6}{9} = {r},
  cell{6}{10} = {r},
  cell{6}{11} = {r},
  cell{6}{12} = {r},
  cell{6}{13} = {r},
  cell{7}{2} = {r},
  cell{7}{3} = {r},
  cell{7}{4} = {r},
  cell{7}{5} = {r},
  cell{7}{6} = {r},
  cell{7}{7} = {r},
  cell{7}{8} = {r},
  cell{7}{9} = {r},
  cell{7}{10} = {r},
  cell{7}{11} = {r},
  cell{7}{12} = {r},
  cell{7}{13} = {r},
  cell{8}{2} = {r},
  cell{8}{3} = {r},
  cell{8}{4} = {r},
  cell{8}{5} = {r},
  cell{8}{6} = {r},
  cell{8}{7} = {r},
  cell{8}{8} = {r},
  cell{8}{9} = {r},
  cell{8}{10} = {r},
  cell{8}{11} = {r},
  cell{8}{12} = {r},
  cell{8}{13} = {r},
  cell{9}{2} = {r},
  cell{9}{3} = {r},
  cell{9}{4} = {r},
  cell{9}{5} = {r},
  cell{9}{6} = {r},
  cell{9}{7} = {r},
  cell{9}{8} = {r},
  cell{9}{9} = {r},
  cell{9}{10} = {r},
  cell{9}{11} = {r},
  cell{9}{12} = {r},
  cell{9}{13} = {r},
  cell{10}{2} = {r},
  cell{10}{3} = {r},
  cell{10}{4} = {r},
  cell{10}{5} = {r},
  cell{10}{6} = {r},
  cell{10}{7} = {r},
  cell{10}{8} = {r},
  cell{10}{9} = {r},
  cell{10}{10} = {r},
  cell{10}{11} = {r},
  cell{10}{12} = {r},
  cell{10}{13} = {r},
  vline{2-3,6,9} = {1}{},
  vline{3-13} = {2}{},
  vline{2-13} = {3-10}{},
  hline{1,3,11} = {-}{},
  hline{2} = {1-13}{},
  vline{2-3} = {1-10}{}, %
  vline{5,8,11} = {1-2}{}, %
}
Models       & CIFAR-10H &       &       & MLRSNet &       &       & FER+   &       &       & ImageNet-Real &       &       \\
             & LAC       & APS   & RAPS  & LAC     & APS   & RAPS  & LAC    & APS   & RAPS  & LAC           & APS   & RAPS  \\
   ResNet18 & 0.131 & 0.231 & 0.231 &  0.045 &  0.064 &  0.064 & 0.365 & 0.325 & 0.326 & 0.320 & 0.316 & 0.316 \\
    ResNet34 & 0.125 & 0.209 & 0.199 &  0.020 &  0.051 &  0.052 & 0.363 & 0.240 & 0.232 & 0.353 & 0.355 & 0.355 \\
    ResNet50 & 0.144 & 0.215 & 0.219 &  0.038 &  0.049 &  0.050 & 0.364 & 0.302 & 0.280 & 0.361 & 0.364 & 0.364 \\
       VGG-16 & 0.137 & 0.230 & 0.220 & -0.010 & -0.016 & -0.025 & 0.409 & 0.277 & 0.268 & 0.344 & 0.347 & 0.347 \\
       VGG-19 & 0.148 & 0.222 & 0.215 &  0.006 &  0.007 & -0.003 & 0.390 & 0.265 & 0.252 & 0.352 & 0.349 & 0.349 \\
 DenseNet121 & 0.018 & 0.221 & 0.204 &  0.037 &  0.058 &  0.058 & 0.393 & 0.177 & 0.168 & 0.351 & 0.357 & 0.357 \\
 DenseNet161 & 0.056 & 0.201 & 0.196 &  0.023 &  0.037 &  0.037 & 0.402 & 0.253 & 0.237 & 0.348 & 0.355 & 0.355 \\
MobileNet-v2 & 0.109 & 0.256 & 0.248 &  0.060 &  0.064 &  0.066 & 0.374 & 0.315 & 0.307 & 0.332 & 0.331 & 0.331 
\end{tblr}
\end{table*}

\begin{table*}[!h]
\caption{Spearman's rank correlation coefficient, $r_s$, $p < .001$, between conformal prediction set sizes and softmax entropy.}
\label{table:spearman_cp_against_softmax_entropy}
\centering
\begin{tblr}{
  row{2} = {r},
  cell{1}{1} = {r=2}{},
  cell{1}{2} = {c=3}{c},
  cell{1}{5} = {c=3}{c},
  cell{1}{8} = {c=3}{c},
  cell{1}{11} = {c=3}{c},
  cell{3}{2} = {r},
  cell{3}{3} = {r},
  cell{3}{4} = {r},
  cell{3}{5} = {r},
  cell{3}{6} = {r},
  cell{3}{7} = {r},
  cell{3}{8} = {r},
  cell{3}{9} = {r},
  cell{3}{10} = {r},
  cell{3}{11} = {r},
  cell{3}{12} = {r},
  cell{3}{13} = {r},
  cell{4}{2} = {r},
  cell{4}{3} = {r},
  cell{4}{4} = {r},
  cell{4}{5} = {r},
  cell{4}{6} = {r},
  cell{4}{7} = {r},
  cell{4}{8} = {r},
  cell{4}{9} = {r},
  cell{4}{10} = {r},
  cell{4}{11} = {r},
  cell{4}{12} = {r},
  cell{4}{13} = {r},
  cell{5}{2} = {r},
  cell{5}{3} = {r},
  cell{5}{4} = {r},
  cell{5}{5} = {r},
  cell{5}{6} = {r},
  cell{5}{7} = {r},
  cell{5}{8} = {r},
  cell{5}{9} = {r},
  cell{5}{10} = {r},
  cell{5}{11} = {r},
  cell{5}{12} = {r},
  cell{5}{13} = {r},
  cell{6}{2} = {r},
  cell{6}{3} = {r},
  cell{6}{4} = {r},
  cell{6}{5} = {r},
  cell{6}{6} = {r},
  cell{6}{7} = {r},
  cell{6}{8} = {r},
  cell{6}{9} = {r},
  cell{6}{10} = {r},
  cell{6}{11} = {r},
  cell{6}{12} = {r},
  cell{6}{13} = {r},
  cell{7}{2} = {r},
  cell{7}{3} = {r},
  cell{7}{4} = {r},
  cell{7}{5} = {r},
  cell{7}{6} = {r},
  cell{7}{7} = {r},
  cell{7}{8} = {r},
  cell{7}{9} = {r},
  cell{7}{10} = {r},
  cell{7}{11} = {r},
  cell{7}{12} = {r},
  cell{7}{13} = {r},
  cell{8}{2} = {r},
  cell{8}{3} = {r},
  cell{8}{4} = {r},
  cell{8}{5} = {r},
  cell{8}{6} = {r},
  cell{8}{7} = {r},
  cell{8}{8} = {r},
  cell{8}{9} = {r},
  cell{8}{10} = {r},
  cell{8}{11} = {r},
  cell{8}{12} = {r},
  cell{8}{13} = {r},
  cell{9}{2} = {r},
  cell{9}{3} = {r},
  cell{9}{4} = {r},
  cell{9}{5} = {r},
  cell{9}{6} = {r},
  cell{9}{7} = {r},
  cell{9}{8} = {r},
  cell{9}{9} = {r},
  cell{9}{10} = {r},
  cell{9}{11} = {r},
  cell{9}{12} = {r},
  cell{9}{13} = {r},
  cell{10}{2} = {r},
  cell{10}{3} = {r},
  cell{10}{4} = {r},
  cell{10}{5} = {r},
  cell{10}{6} = {r},
  cell{10}{7} = {r},
  cell{10}{8} = {r},
  cell{10}{9} = {r},
  cell{10}{10} = {r},
  cell{10}{11} = {r},
  cell{10}{12} = {r},
  cell{10}{13} = {r},
  vline{2-3,6,9} = {1}{},
  vline{3-13} = {2}{},
  vline{2-13} = {3-10}{},
  hline{1,3,11} = {-}{},
  hline{2} = {1-13}{},
  vline{2-3} = {1-10}{}, %
  vline{5,8,11} = {1-2}{}, %
}
Models       & CIFAR-10H &       &       & MLRSNet &       &       & FER+   &       &       & ImageNet-Real &       &       \\
             & LAC       & APS   & RAPS  & LAC     & APS   & RAPS  & LAC    & APS   & RAPS  & LAC           & APS   & RAPS  \\
ResNet18 & 0.326 & 0.614 & 0.614 & 0.432 & 0.723 & 0.733 & 0.824 & 0.935 & 0.934 & 0.924 & 0.953 & 0.953 \\
    ResNet34 & 0.391 & 0.595 & 0.573 & 0.461 & 0.706 & 0.712 & 0.785 & 0.918 & 0.916 & 0.881 & 0.932 & 0.933 \\
    ResNet50 & 0.359 & 0.547 & 0.558 & 0.505 & 0.719 & 0.720 & 0.804 & 0.912 & 0.903 & 0.828 & 0.909 & 0.910 \\
       VGG-16 & 0.361 & 0.583 & 0.561 & 0.731 & 0.746 & 0.776 & 0.804 & 0.902 & 0.898 & 0.898 & 0.943 & 0.942 \\
       VGG-19 & 0.324 & 0.551 & 0.519 & 0.814 & 0.820 & 0.860 & 0.800 & 0.917 & 0.911 & 0.892 & 0.940 & 0.939 \\
 DenseNet121 & 0.091 & 0.571 & 0.523 & 0.378 & 0.675 & 0.683 & 0.790 & 0.885 & 0.877 & 0.862 & 0.932 & 0.932 \\
 DenseNet161 & 0.157 & 0.536 & 0.521 & 0.253 & 0.615 & 0.613 & 0.790 & 0.876 & 0.863 & 0.814 & 0.898 & 0.897 \\
MobileNet-v2 & 0.283 & 0.614 & 0.593 & 0.819 & 0.892 & 0.903 & 0.828 & 0.930 & 0.930 & 0.903 & 0.943 & 0.944
\end{tblr}
\end{table*}

\subsection{Similarity analysis}


All methods achieve low subset-accuracy and relatively higher Hamming loss, showing the substantial difference between prediction sets and human annotations. Precision and recall are influenced by prediction set size and the number of distinct human labels. Smaller prediction sets and more diverse human annotations, such as CIFAR-10H, MLRSNet, and FER+, yield higher precision. In contrast, ImageNet-ReaL exhibits lower precision and higher recall due to fewer human labels (five annotators chose from eight labels). Tabular results of our similarity analysis are presented in \Cref{table:similarity_annotation_vs_cp_cifar10h,table:similarity_annotation_vs_cp_mlrsnet,table:similarity_annotation_vs_cp_ferplus,table:similarity_annotation_vs_cp_imagenetreal} in the Supplementary material. 



\subsection{Expected Calibration Error}


ECE of all the models is presented in Table~\ref{table:ece} in Supplementary material. The majority of models exhibit strong calibration with a median ECE of 0.024 across all experiments. The VGG-16 and VGG-19 models show the highest degree of miscalibration on the MLRSNet dataset.


\subsection{Conformal predictors' performance}

We report the models' accuracy and the conformal predictors' coverage, SSC, and mean set size ($\bar{w}$). The APS and RAPS methods consistently outperform the LAC method in covering the true class at the cost of a larger mean set size. Accuracy and coverage of all the models and conformal predictors on the CIFAR-10H dataset are shown in Table~\ref{table:coverage_only_performance_cifar10h}. A summary of coverage, SSC, and $\bar{w}$ for all the models is reported in Tables~\ref{table:avg_performance_coverage}-\ref{table:avg_performance_w}. Detailed tabular report of the models' and conformal predictors' performance on all four datasets can be found in Tables~\ref{table:coverage_performance_cifar10h}-\ref{table:coverage_performance_imagenet-real} in Supplementary material.

Notably, VGG-16 and VGG-19, which show high miscalibration on the MLRSNet dataset, also achieve lower accuracy (See Table~\ref{table:coverage_performance_mlrsnet} in Supplementary material). Although their coverage under the LAC method is comparable to that of other models, they are consistently outperformed when using the APS and RAPS methods.

While this trend is less apparent on datasets with similar model accuracies, we observe that lower-accuracy models consistently yield larger mean prediction set sizes across all methods. This pattern is evident for VGG-16, VGG-19, and MobileNet-v2 on the MLRSNet dataset (See \Cref{table:coverage_performance_mlrsnet} in Supplementary material), and similarly for ResNet18, VGG-16, VGG-19, and MobileNet-v2 on the ImageNet-ReaL dataset (See \Cref{table:coverage_performance_imagenet-real} in Supplementary material).

LAC frequently generates smaller set sizes. This can be seen summarized in Table~\ref{table:avg_performance_w} (See details in Tables~\ref{table:coverage_performance_cifar10h}-\ref{table:coverage_performance_imagenet-real} in Supplementary material). LAC can also lead to empty prediction sets, as is especially the case for the models trained on the CIFAR-10H, MLRSNet, and ImageNet-Real datasets (See the distribution of the prediction set sizes in Figure~\ref{figure:prediction_set_sizes_distribution} in the Supplementary material). 

\begin{figure*}[h]
\centering
\includegraphics[width=0.95\textwidth]{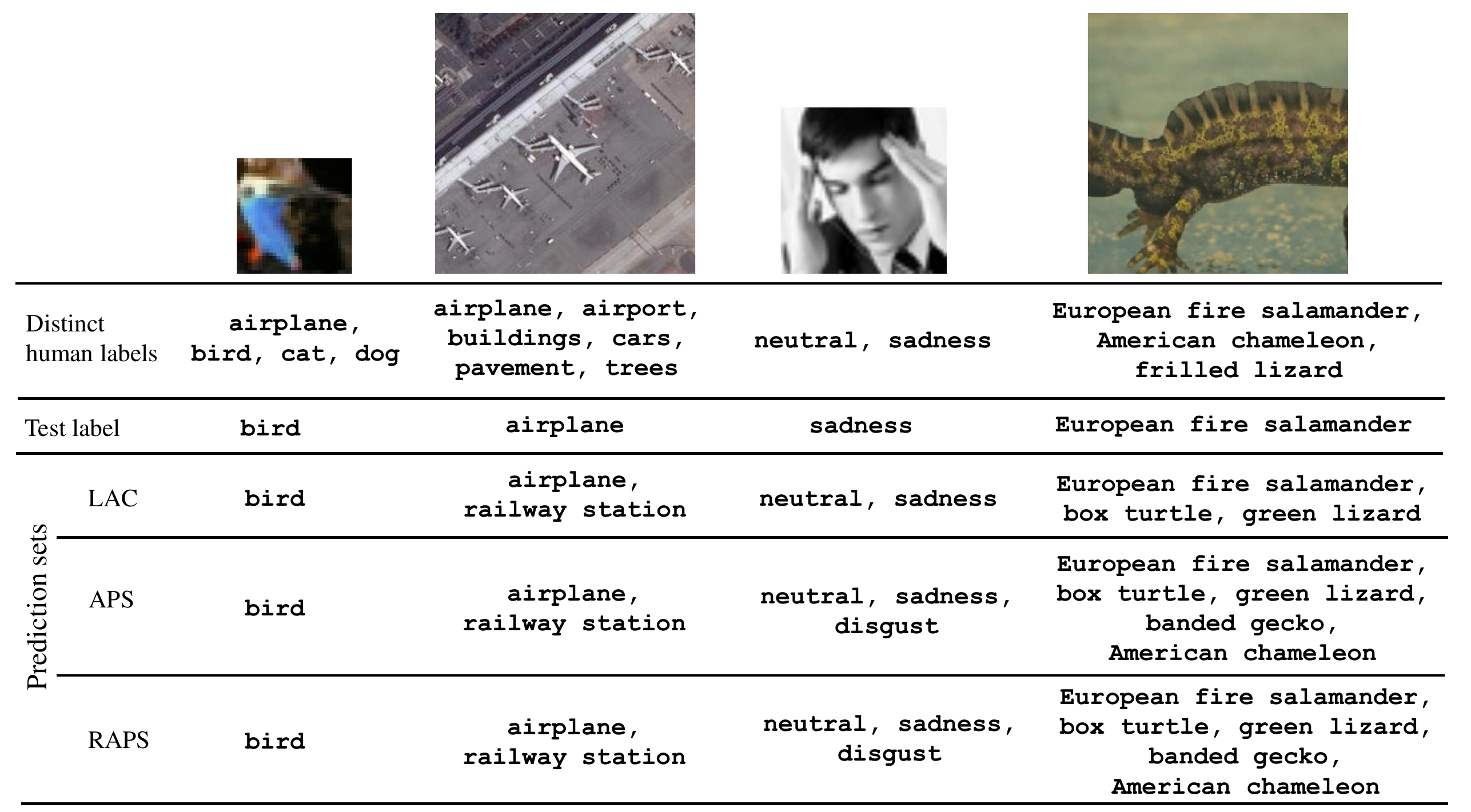}
\caption{Conformal prediction set outputs for sample images from the four datasets. Columns from left to right: CIFAR-10H (leftmost), MLRSNet, FER+, and ImageNet-Real (rightmost).}
\label{figure:sample_cp_outputs}
\end{figure*}

Conformal prediction sets constructed with the ResNet18 model for sample images from all datasets are shown in Figure~\ref{figure:sample_cp_outputs}. As shown in the figure, while all approaches successfully cover the true class, their performance at capturing the ambiguity in human annotations varies and is usually limited.


\begin{table}
\centering
\caption{Coverage of conformal predictors' on the CIFAR-10H dataset.}
\label{table:coverage_only_performance_cifar10h}
\begin{tblr}{
  row{2} = {r},
  cell{1}{1} = {r=2}{},
  cell{1}{2} = {r=2}{},
  cell{1}{3} = {c=3}{c},
  cell{3}{2} = {r},
  cell{3}{3} = {r},
  cell{3}{4} = {r},
  cell{3}{5} = {r},
  cell{4}{2} = {r},
  cell{4}{3} = {r},
  cell{4}{4} = {r},
  cell{4}{5} = {r},
  cell{5}{2} = {r},
  cell{5}{3} = {r},
  cell{5}{4} = {r},
  cell{5}{5} = {r},
  cell{6}{2} = {r},
  cell{6}{3} = {r},
  cell{6}{4} = {r},
  cell{6}{5} = {r},
  cell{7}{2} = {r},
  cell{7}{3} = {r},
  cell{7}{4} = {r},
  cell{7}{5} = {r},
  cell{8}{2} = {r},
  cell{8}{3} = {r},
  cell{8}{4} = {r},
  cell{8}{5} = {r},
  cell{9}{2} = {r},
  cell{9}{3} = {r},
  cell{9}{4} = {r},
  cell{9}{5} = {r},
  cell{10}{2} = {r},
  cell{10}{3} = {r},
  cell{10}{4} = {r},
  cell{10}{5} = {r},
  vline{2-3} = {1-3}{},
  vline{4-5} = {2-3}{},
  vline{2-5} = {3-10}{},
  hline{1,3,11} = {-}{},
  hline{2} = {3-5}{},
}
Models       & Accuracy & Coverage &       &       \\
             &          & LAC      & APS   & RAPS  \\
ResNet18     & 0.930    & 0.944    & 0.977 & 0.978 \\
ResNet34     & 0.933    & 0.953    & 0.979 & 0.976 \\
ResNet50     & 0.936    & 0.953    & 0.974 & 0.975 \\
VGG-16       & 0.941    & 0.955    & 0.982 & 0.979 \\
VGG-19       & 0.939    & 0.951    & 0.976 & 0.973 \\
DenseNet121  & 0.940    & 0.942    & 0.979 & 0.975 \\
DenseNet161  & 0.940    & 0.944    & 0.977 & 0.976 \\
MobileNet-v2 & 0.939    & 0.951    & 0.984 & 0.982 
\end{tblr}
\end{table}

\begin{table*}
\centering
\caption{Mean ($\pm$ std) of models' coverage for each dataset.}
\label{table:avg_performance_coverage}
\begin{tblr}{
  row{2} = {r},
  column{2} = {r},
  cell{1}{1} = {r=2}{},
  cell{1}{2} = {r=2}{},
  cell{1}{3} = {c=3}{c},
  cell{3}{3} = {r},
  cell{3}{4} = {r},
  cell{3}{5} = {r},
  cell{4}{3} = {r},
  cell{4}{4} = {r},
  cell{4}{5} = {r},
  cell{5}{3} = {r},
  cell{5}{4} = {r},
  cell{5}{5} = {r},
  cell{6}{3} = {r},
  cell{6}{4} = {r},
  cell{6}{5} = {r},
  vline{2-3} = {1-6}{},
  vline{4-5} = {2-6}{},
  vline{2-5} = {3-6}{},
  hline{1,3,7} = {-}{},
  hline{2} = {3-5}{},
}
Dataset       & Accuracy          & Coverage          &                   &                   \\
              &                   & LAC               & APS               & RAPS              \\
CIFAR-10H     & 0.937 $\pm$ 0.004 & 0.949 $\pm$ 0.005 & 0.978 $\pm$ 0.003 & 0.977 $\pm$ 0.003 \\
MLRSNet       & 0.903 $\pm$ 0.038 & 0.952 $\pm$ 0.002 & 0.976 $\pm$ 0.012 & 0.979 $\pm$ 0.007 \\
FER+          & 0.832 $\pm$ 0.006 & 0.943 $\pm$ 0.002 & 0.963 $\pm$ 0.004 & 0.964 $\pm$ 0.004 \\
ImageNet-ReaL & 0.759 $\pm$ 0.023 & 0.902 $\pm$ 0.001 & 0.943 $\pm$ 0.003 & 0.943 $\pm$ 0.003 
\end{tblr}
\end{table*}

\begin{table*}
\centering
\caption{Mean ($\pm$ std) of models' SSC for each dataset.}
\label{table:avg_performance_acc_ssc}
\begin{tblr}{
  row{2} = {r},
  column{2} = {r},
  cell{1}{1} = {r=2}{},
  cell{1}{2} = {r=2}{},
  cell{1}{3} = {c=3}{c},
  cell{3}{3} = {r},
  cell{3}{4} = {r},
  cell{3}{5} = {r},
  cell{4}{3} = {r},
  cell{4}{4} = {r},
  cell{4}{5} = {r},
  cell{5}{3} = {r},
  cell{5}{4} = {r},
  cell{5}{5} = {r},
  cell{6}{3} = {r},
  cell{6}{4} = {r},
  cell{6}{5} = {r},
  vline{2-3} = {1-6}{},
  vline{4-5} = {2-6}{},
  vline{2-5} = {3-6}{},
  hline{1,3,7} = {-}{},
  hline{2} = {3-5}{},
}
Dataset       & Accuracy          & SSC               &                   &                   \\
              &                   & LAC               & APS               & RAPS              \\
CIFAR-10H     & 0.937 $\pm$ 0.004 & 0.385 $\pm$ 0.428 & 0.939 $\pm$ 0.016 & 0.935 $\pm$ 0.031 \\
MLRSNet       & 0.903 $\pm$ 0.038 & 0.500 $\pm$ 0.416 & 0.704 $\pm$ 0.321 & 0.810 $\pm$ 0.158 \\
FER+          & 0.832 $\pm$ 0.006 & 0.891 $\pm$ 0.051 & 0.900 $\pm$ 0.058 & 0.915 $\pm$ 0.038 \\
ImageNet-ReaL & 0.759 $\pm$ 0.023 & 0.234 $\pm$ 0.327 & 0.000 $\pm$ 0.000 & 0.062 $\pm$ 0.177 \\
\end{tblr}
\end{table*}

\begin{table*}[!h]
\centering
\caption{Mean ($\pm$ std) of models' mean width $\bar{w}$ for each dataset.}
\label{table:avg_performance_w}
\begin{tblr}{
  row{2} = {r},
  cell{1}{1} = {r=2}{},
  cell{1}{2} = {c=3}{c},
  cell{3}{2} = {r},
  cell{3}{3} = {r},
  cell{3}{4} = {r},
  cell{4}{2} = {r},
  cell{4}{3} = {r},
  cell{4}{4} = {r},
  cell{5}{2} = {r},
  cell{5}{3} = {r},
  cell{5}{4} = {r},
  cell{6}{2} = {r},
  cell{6}{3} = {r},
  cell{6}{4} = {r},
  vline{2-4} = {1-6}{},
  hline{1,3,7} = {-}{},
  hline{2} = {2-4}{},
}
Dataset       & $\bar{w}$        &                   &                   \\
              & LAC               & APS               & RAPS              \\
CIFAR-10H     & 1.036 $\pm$ 0.019 & 1.344 $\pm$ 0.058 & 1.297 $\pm$ 0.053 \\
MLRSNet       & 1.340 $\pm$ 0.398 & 1.770 $\pm$ 0.459 & 1.978 $\pm$ 0.750 \\
FER+          & 1.526 $\pm$ 0.033 & 2.188 $\pm$ 0.086 & 2.252 $\pm$ 0.073 \\
ImageNet-ReaL & 2.236 $\pm$ 0.438 & 9.379 $\pm$ 2.035 & 9.412 $\pm$ 2.016 \\
\end{tblr}
\end{table*}

\subsection{Coverage at different prediction set sizes}


Larger prediction sets do not always lead to improved coverage (We show how coverage changes with increasing set sizes for all the datasets and CP methods in \Cref{figure:coverage_against_prediction_set_sizes} in Supplementary material). This is especially the case for the datasets with a large number of categories, MLRSNet (46 categories) and ImageNet-Real (1000 categories). However, some of the models trained on the CIFAR-10H and FER+ datasets led to better coverage with increasing set size.


\section{Discussion}
\label{sec:discussion}

We evaluate the common assumption that conformal predictors capture aleatoric uncertainty. We perform this by measuring their correlation and similarity with the ambiguity present in human annotations across four datasets. Our experiments use a diverse set of datasets with varying degrees of class overlap from human annotators. Overall, we found that conformal predictors find it challenging to capture the class overlap recorded using multiple human annotators. We also observe similar results when considering entropy in human annotations.

In agreement with previous works \cite{angelopoulosuncertainty}, we see that the conformal predictors lead to an increased coverage of the true class for all the datasets and models. This is presented in Table~\ref{table:avg_performance_coverage} and in Tables~\ref{table:coverage_performance_cifar10h}-\ref{table:coverage_performance_imagenet-real} in the Supplementary material. 

Even though conformal predictors often show a strong correlation with aleatoric uncertainty quantifiers such as the softmax entropy (See Table~\ref{table:spearman_cp_against_softmax_entropy}), they usually fail to capture the inherent class overlap and entropy in human annotations in the employed datasets, and show substantial differences from human annotations. For the challenging dataset, MLRSNet, where 97.2\% of the images have overlapping classes (or received more than two distinct human labels), the Spearman's $r_s$ between the prediction set size and class overlap was always very weak. This highlights the particular shortcomings of prediction sets when presented with challenging tasks. We observe weak and a few moderate correlations in cases where class overlap is low, as in the ImageNet-Real dataset (there is class overlap in only 15.8\% of the total images from the ImageNet-ReaL dataset) or where the number of categories is small as in the FER+ dataset (the FER+ has only seven categories, which is the lowest number of categories relative to the rest of the datasets).


In this paper, we focus on aleatoric uncertainty as captured by class overlap. We also quantify aleatoric uncertainty using human annotation entropy whenever such information is available. This form of uncertainty is irreducible because it arises from the nature of the datasets and the label space itself. We use multiple human annotators per instance and define class overlap as the number of distinct labels assigned to that instance; more distinct labels indicate greater intrinsic ambiguity. For example, it is expected that a satellite image of an airport can contain airplanes. So, if we ask several annotators for a single label, we may reasonably get the labels \emph{airport} and \emph{airplane} for the same image. Similar patterns occur in subjective tasks such as emotion recognition, where different but defensible choices can yield multiple distinct labels for the same face. By grounding our evaluation in these naturally occurring ambiguities, we can directly assess whether conformal predictors capture the intrinsic uncertainty present in real data.

By contrast, epistemic uncertainty reflects limitations of the model or data and is, in principle, reducible (e.g., labeling error). To minimize any potential leakage from epistemic factors, we (i) use multiple annotators per item with consistent instructions, (ii) assess ECE and make sure that the trained models have low calibration error, and (iii) rely on datasets and training protocols that yield strong performance in prior work and reproduce these results in our experiments. 




\section{Conclusion}
\label{sec:conclusion}

This work provides the first empirical evaluation of conformal predictors' ability at quantifying aleatoric uncertainty arising from class overlap. To probe this capability, we used fitting datasets whose inherent ambiguity was identified by collecting labels from multiple annotators (ranging between five and fifty participants) per single instance. We applied three conformal prediction methods across four datasets exhibiting varying degrees of class overlap. Our findings indicate that the Spearman's $r_s$ between conformal prediction set uncertainty, measured by prediction set size, and class overlap, measured by the number of distinct human labels, was frequently in the very weak to weak range. These results suggest that conformal predictors, while offering valid coverage, often fail to reflect the underlying ambiguity in data caused by class overlap. We also observed that the correlation between prediction set size and class overlap improves with the prevalence of larger prediction sets. While we hypothesized that increasing set size would consistently enhance coverage or the likelihood of including the true class, this expectation did not hold across most of our experiments. These findings underscore that prediction set size should not be uncritically interpreted as a proxy for aleatoric uncertainty, particularly in the presence of inherent labeling ambiguity. This paper focuses on aleatoric uncertainty arising from class overlap. Although other sources of aleatoric uncertainty, such as intra-class variability or label noise, are often intertwined with class overlap, they are not explicitly addressed in this work and are left for future investigation.



\section*{Acknowledgments}

This work was partially supported by the Wallenberg AI, Autonomous Systems and Software Program (WASP) funded by the Knut and Alice Wallenberg Foundation.



{
    \small
    \bibliographystyle{ieeenat_fullname}
    \bibliography{main}
}

\clearpage

\appendix
\renewcommand{\thefigure}{A\arabic{figure}}
\renewcommand{\thetable}{A\arabic{table}}
\setcounter{figure}{0}
\setcounter{table}{0}
\pagenumbering{Roman} 
\setcounter{page}{1}

\section*{Supplementary material}


\section*{Paper \#1305 : Performance of Conformal Prediction in Capturing Aleatoric Uncertainty}

\paragraph{Models' and conformal predictors' performance}

The models' accuracy and the conformal predictors' coverage, SSC, and mean set size ($\bar{w}$) on all four datasets are presented in Table~\ref{table:coverage_performance_cifar10h} to Table~\ref{table:coverage_performance_imagenet-real}.

\paragraph{Distribution of prediction set sizes}

\begin{figure*}
\centering
\includegraphics[width=1\textwidth]{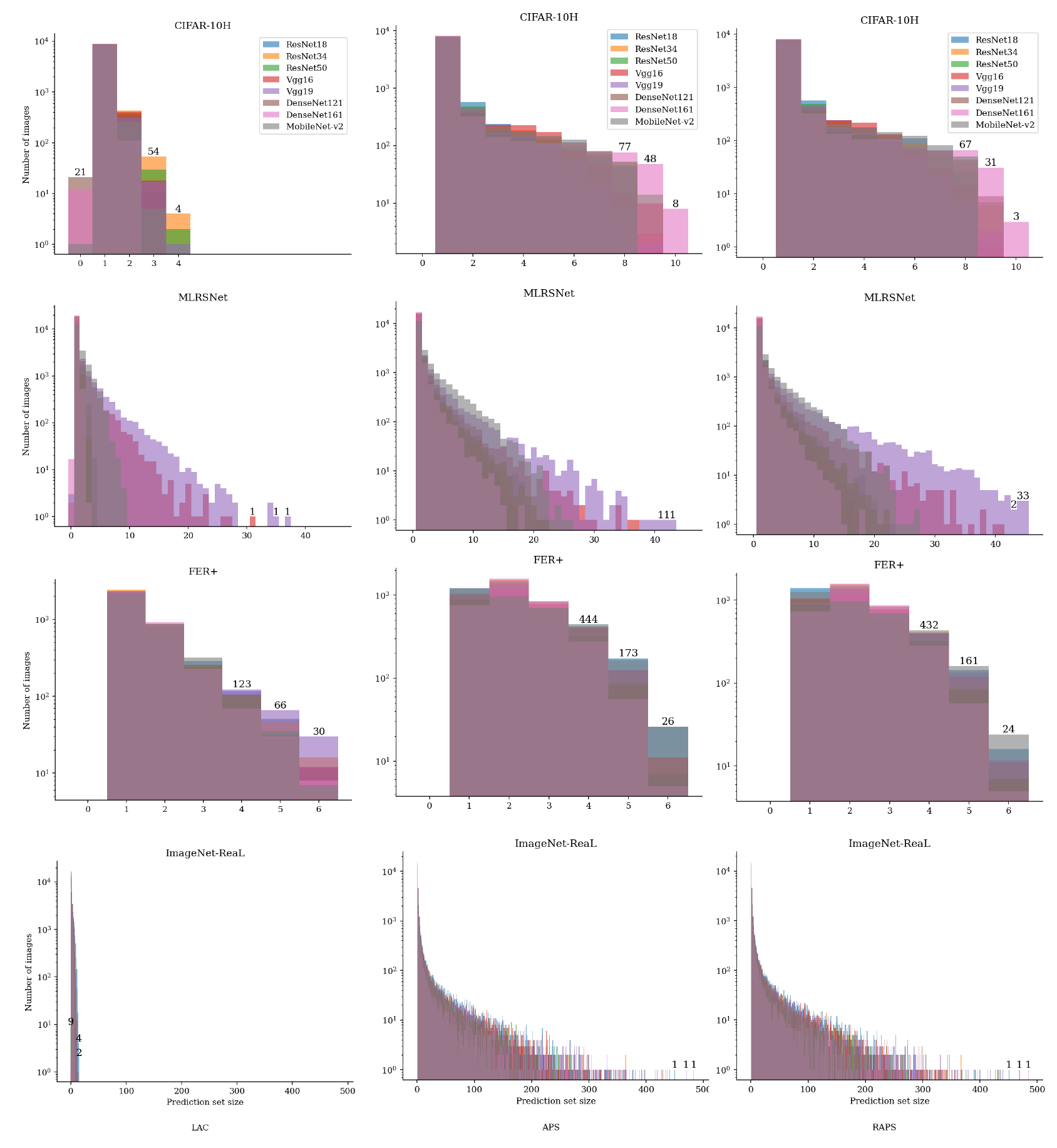}
\caption{Distribution of prediction set sizes. The y-axis is log-scaled for easier visualization of the imbalanced counts. The count of the largest three prediction set sizes is shown on top of the bars. The three columns, from left to right, represent LAC, APS, and RAPS, respectively.}
\label{figure:prediction_set_sizes_distribution}
\end{figure*}

Figure~\ref{figure:prediction_set_sizes_distribution} shows the prediction set size distribution of the three conformal predictors.

\paragraph{Coverage at different prediction set sizes}

\begin{figure*}[h]
\centering
\includegraphics[width=1\textwidth]{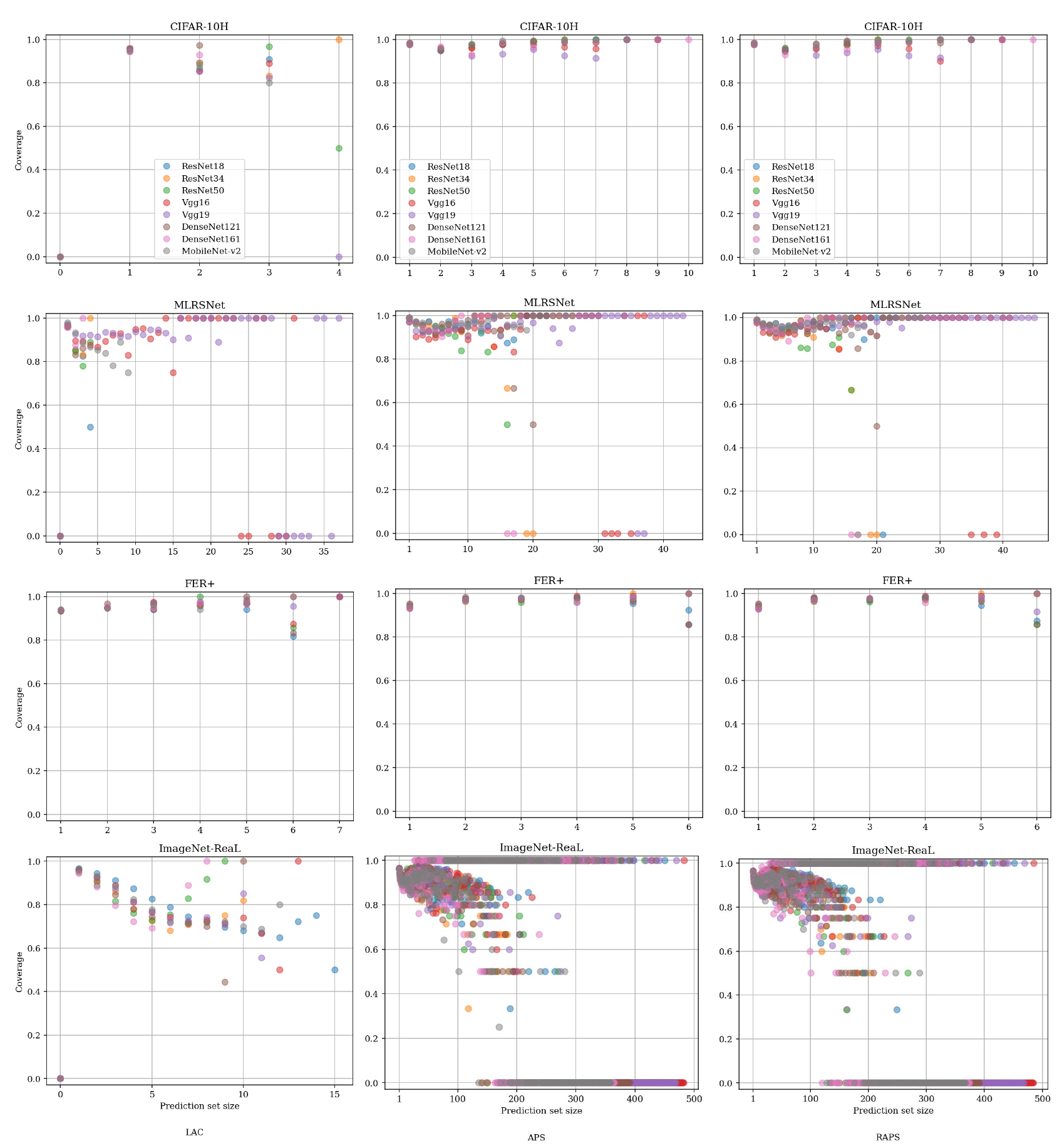}
\caption{Coverage at different prediction set sizes. The X-axis shows prediction set sizes, and the Y-axis shows the coverage. The three columns, from left to right, represent LAC, APS, and RAPS, respectively.}
\label{figure:coverage_against_prediction_set_sizes}
\end{figure*}

We show how coverage changes with increasing set sizes in \Cref{figure:coverage_against_prediction_set_sizes}.

\paragraph{Impact of prevalence of larger prediction set sizes on correlation analysis}

\begin{figure*}
\centering
\includegraphics[width=1\textwidth]{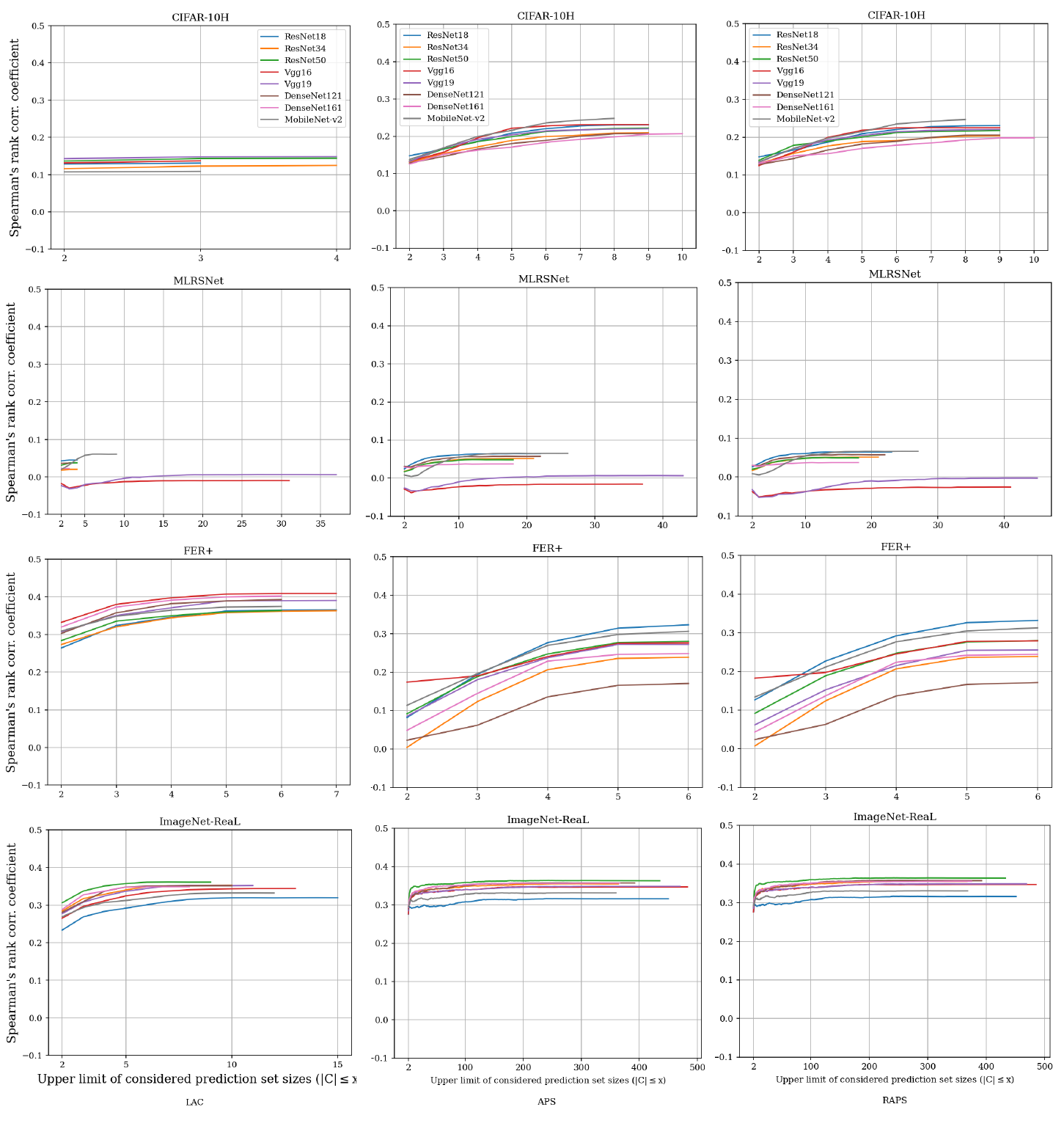}
\caption{Spearman's rank correlation coefficient, $r_s$, $p < .001$, between prediction set sizes and class overlap with an increased prevalence of larger prediction sets. The X-axis shows the upper limit of the considered prediction set sizes $(|C| \leq x)$, and the Y-axis shows Spearman's rank correlation coefficient. The three columns, from left to right, represent LAC, APS, and RAPS, respectively.}
\label{figure:spearman_correlation_against_prediction_set_sizes}
\end{figure*}

We show how the correlation between prediction set sizes and class overlap changes with the prevalence of larger set sizes in Figure~\ref{figure:spearman_correlation_against_prediction_set_sizes}. We follow an incremental approach to see the effect of larger set sizes on the correlation analysis. We start with prediction sets with size $\leq$ 2 and incrementally add sets with larger sizes. The Spearman's rank correlation coefficient of prediction sets that contain larger set sizes shows improvement, albeit small.

\paragraph{Similarity between conformal prediction sets and human annotation}

We assess the similarity between the prediction sets of the conformal predictors and human annotations using the metrics precision, recall, subset-accuracy (S. Acc), and Hamming loss. The results on all the datasets are presented in \Cref{table:similarity_annotation_vs_cp_cifar10h,table:similarity_annotation_vs_cp_mlrsnet,table:similarity_annotation_vs_cp_ferplus,table:similarity_annotation_vs_cp_imagenetreal}.

\paragraph{Correlation between conformal prediction sets and human annotation entropy}

We present the results of the correlation analysis between prediction sets and human annotation entropy on the CIFAR-10H and FER+ datasets in \Cref{table:spearman_cp_sets_and_human_annotation_entropy}.

\paragraph{Expected Calibration Error}

Table~\ref{table:ece} presents the Expected Calibration Error (ECE) using M=15 bins for all the models.

\clearpage

\begin{table*}[!h]
\centering
\caption{Models' and conformal predictors' performance on the CIFAR-10H dataset.}
\label{table:coverage_performance_cifar10h}
\begin{tblr}{
  row{2} = {r},
  cell{1}{1} = {r=2}{},
  cell{1}{2} = {r=2}{},
  cell{1}{3} = {c=3}{c},
  cell{1}{6} = {c=3}{c},
  cell{1}{9} = {c=3}{c},
  cell{3}{2} = {r},
  cell{3}{3} = {r},
  cell{3}{4} = {r},
  cell{3}{5} = {r},
  cell{3}{6} = {r},
  cell{3}{7} = {r},
  cell{3}{8} = {r},
  cell{3}{9} = {r},
  cell{3}{10} = {r},
  cell{3}{11} = {r},
  cell{4}{2} = {r},
  cell{4}{3} = {r},
  cell{4}{4} = {r},
  cell{4}{5} = {r},
  cell{4}{6} = {r},
  cell{4}{7} = {r},
  cell{4}{8} = {r},
  cell{4}{9} = {r},
  cell{4}{10} = {r},
  cell{4}{11} = {r},
  cell{5}{2} = {r},
  cell{5}{3} = {r},
  cell{5}{4} = {r},
  cell{5}{5} = {r},
  cell{5}{6} = {r},
  cell{5}{7} = {r},
  cell{5}{8} = {r},
  cell{5}{9} = {r},
  cell{5}{10} = {r},
  cell{5}{11} = {r},
  cell{6}{2} = {r},
  cell{6}{3} = {r},
  cell{6}{4} = {r},
  cell{6}{5} = {r},
  cell{6}{6} = {r},
  cell{6}{7} = {r},
  cell{6}{8} = {r},
  cell{6}{9} = {r},
  cell{6}{10} = {r},
  cell{6}{11} = {r},
  cell{7}{2} = {r},
  cell{7}{3} = {r},
  cell{7}{4} = {r},
  cell{7}{5} = {r},
  cell{7}{6} = {r},
  cell{7}{7} = {r},
  cell{7}{8} = {r},
  cell{7}{9} = {r},
  cell{7}{10} = {r},
  cell{7}{11} = {r},
  cell{8}{2} = {r},
  cell{8}{3} = {r},
  cell{8}{4} = {r},
  cell{8}{5} = {r},
  cell{8}{6} = {r},
  cell{8}{7} = {r},
  cell{8}{8} = {r},
  cell{8}{9} = {r},
  cell{8}{10} = {r},
  cell{8}{11} = {r},
  cell{9}{2} = {r},
  cell{9}{3} = {r},
  cell{9}{4} = {r},
  cell{9}{5} = {r},
  cell{9}{6} = {r},
  cell{9}{7} = {r},
  cell{9}{8} = {r},
  cell{9}{9} = {r},
  cell{9}{10} = {r},
  cell{9}{11} = {r},
  cell{10}{2} = {r},
  cell{10}{3} = {r},
  cell{10}{4} = {r},
  cell{10}{5} = {r},
  cell{10}{6} = {r},
  cell{10}{7} = {r},
  cell{10}{8} = {r},
  cell{10}{9} = {r},
  cell{10}{10} = {r},
  cell{10}{11} = {r},
  vline{2-3} = {1-10}{},
  vline{4-11} = {2}{},
  vline{6}   = {1-2}{},
  vline{9}   = {1-2}{},
  vline{2-11} = {3-10}{},
  hline{1,3,11} = {-}{},
  hline{2} = {3-11}{},
}
Models        &  Accuracy & LAC      &       &            & APS      &       &            & RAPS     &       &            \\

              &           & Coverage & SSC   & $\bar{w}$ & Coverage & SSC   & $\bar{w}$ & Coverage & SSC   & $\bar{w}$ \\
    ResNet18 &     0.930 &     0.944 & 0.894 &       1.040 &     0.977 & 0.953 &       1.357 &     0.978 & 0.952 &       1.361 \\
    ResNet34 &     0.933 &     0.953 & 0.833 &       1.062 &     0.979 & 0.953 &       1.369 &     0.976 & 0.950 &       1.304 \\
    ResNet50 &     0.936 &     0.953 & 0.500 &       1.050 &     0.974 & 0.939 &       1.263 &     0.975 & 0.959 &       1.285 \\
       VGG-16 &     0.941 &     0.955 & 0.853 &       1.049 &     0.982 & 0.930 &       1.328 &     0.979 & 0.882 &       1.263 \\
       VGG-19 &     0.939 &     0.951 & 0.000 &       1.039 &     0.976 & 0.907 &       1.264 &     0.973 & 0.895 &       1.207 \\
 DenseNet121 &     0.940 &     0.942 & 0.000 &       1.006 &     0.979 & 0.938 &       1.395 &     0.975 & 0.947 &       1.274 \\
 DenseNet161 &     0.940 &     0.944 & 0.000 &       1.011 &     0.977 & 0.941 &       1.347 &     0.976 & 0.927 &       1.309 \\
MobileNet-v2 &     0.939 &     0.951 & 0.000 &       1.030 &     0.984 & 0.953 &       1.425 &     0.982 & 0.966 &       1.370     
\end{tblr}
\end{table*}

\begin{table*}[!h]
\centering
\caption{Models' and conformal predictors' performance on the MLRSNet dataset.}
\label{table:coverage_performance_mlrsnet}
\begin{tblr}{
  row{2} = {r},
  cell{1}{1} = {r=2}{},
  cell{1}{2} = {r=2}{},
  cell{1}{3} = {c=3}{c},
  cell{1}{6} = {c=3}{c},
  cell{1}{9} = {c=3}{c},
  cell{3}{2} = {r},
  cell{3}{3} = {r},
  cell{3}{4} = {r},
  cell{3}{5} = {r},
  cell{3}{6} = {r},
  cell{3}{7} = {r},
  cell{3}{8} = {r},
  cell{3}{9} = {r},
  cell{3}{10} = {r},
  cell{3}{11} = {r},
  cell{4}{2} = {r},
  cell{4}{3} = {r},
  cell{4}{4} = {r},
  cell{4}{5} = {r},
  cell{4}{6} = {r},
  cell{4}{7} = {r},
  cell{4}{8} = {r},
  cell{4}{9} = {r},
  cell{4}{10} = {r},
  cell{4}{11} = {r},
  cell{5}{2} = {r},
  cell{5}{3} = {r},
  cell{5}{4} = {r},
  cell{5}{5} = {r},
  cell{5}{6} = {r},
  cell{5}{7} = {r},
  cell{5}{8} = {r},
  cell{5}{9} = {r},
  cell{5}{10} = {r},
  cell{5}{11} = {r},
  cell{6}{2} = {r},
  cell{6}{3} = {r},
  cell{6}{4} = {r},
  cell{6}{5} = {r},
  cell{6}{6} = {r},
  cell{6}{7} = {r},
  cell{6}{8} = {r},
  cell{6}{9} = {r},
  cell{6}{10} = {r},
  cell{6}{11} = {r},
  cell{7}{2} = {r},
  cell{7}{3} = {r},
  cell{7}{4} = {r},
  cell{7}{5} = {r},
  cell{7}{6} = {r},
  cell{7}{7} = {r},
  cell{7}{8} = {r},
  cell{7}{9} = {r},
  cell{7}{10} = {r},
  cell{7}{11} = {r},
  cell{8}{2} = {r},
  cell{8}{3} = {r},
  cell{8}{4} = {r},
  cell{8}{5} = {r},
  cell{8}{6} = {r},
  cell{8}{7} = {r},
  cell{8}{8} = {r},
  cell{8}{9} = {r},
  cell{8}{10} = {r},
  cell{8}{11} = {r},
  cell{9}{2} = {r},
  cell{9}{3} = {r},
  cell{9}{4} = {r},
  cell{9}{5} = {r},
  cell{9}{6} = {r},
  cell{9}{7} = {r},
  cell{9}{8} = {r},
  cell{9}{9} = {r},
  cell{9}{10} = {r},
  cell{9}{11} = {r},
  cell{10}{2} = {r},
  cell{10}{3} = {r},
  cell{10}{4} = {r},
  cell{10}{5} = {r},
  cell{10}{6} = {r},
  cell{10}{7} = {r},
  cell{10}{8} = {r},
  cell{10}{9} = {r},
  cell{10}{10} = {r},
  cell{10}{11} = {r},
  vline{2-3} = {1-10}{},
  vline{4-11} = {2}{},
  vline{2-11} = {3-10}{},
  vline{6}   = {1-2}{},
  vline{9}   = {1-2}{},
  hline{1,3,11} = {-}{},
  hline{2} = {3-11}{},
}
Models        &  Accuracy & LAC      &       &            & APS      &       &            & RAPS     &       &            \\
              &           & Coverage & SSC   & $\bar{w}$ & Coverage & SSC   & $\bar{w}$ & Coverage & SSC   & $\bar{w}$ \\
    ResNet18 &     0.927 &     0.951 & 0.000 &       1.081 &     0.984 & 0.889 &       1.575 &     0.985 & 0.900 &       1.624 \\
    ResNet34 &     0.923 &     0.951 & 0.831 &       1.090 &     0.982 & 0.500 &       1.500 &     0.982 & 0.750 &       1.526 \\
    ResNet50 &     0.918 &     0.949 & 0.780 &       1.113 &     0.978 & 0.667 &       1.504 &     0.978 & 0.667 &       1.508 \\
       VGG-16 &     0.872 &     0.950 & 0.750 &       1.613 &     0.955 & 0.833 &       1.798 &     0.963 & 0.887 &       2.059 \\
       VGG-19 &     0.844 &     0.955 & 0.889 &       2.106 &     0.958 & 0.875 &       2.352 &     0.974 & 0.953 &       3.386 \\
 DenseNet121 &     0.933 &     0.951 & 0.000 &       1.059 &     0.984 & 0.000 &       1.513 &     0.985 & 0.500 &       1.547 \\
 DenseNet161 &     0.946 &     0.956 & 0.000 &       1.026 &     0.985 & 0.928 &       1.320 &     0.984 & 0.905 &       1.316 \\
MobileNet-v2 &     0.860 &     0.953 & 0.750 &       1.635 &     0.978 & 0.937 &       2.596 &     0.981 & 0.917 &       2.859
\end{tblr}
\end{table*}

\begin{table*}[!h]
\centering
\caption{Models' and conformal predictors' performance on the FER+ dataset.}
\label{table:coverage_performance_ferplus}
\begin{tblr}{
  row{2} = {r},
  cell{1}{1} = {r=2}{},
  cell{1}{2} = {r=2}{},
  cell{1}{3} = {c=3}{c},
  cell{1}{6} = {c=3}{c},
  cell{1}{9} = {c=3}{c},
  cell{3}{2} = {r},
  cell{3}{3} = {r},
  cell{3}{4} = {r},
  cell{3}{5} = {r},
  cell{3}{6} = {r},
  cell{3}{7} = {r},
  cell{3}{8} = {r},
  cell{3}{9} = {r},
  cell{3}{10} = {r},
  cell{3}{11} = {r},
  cell{4}{2} = {r},
  cell{4}{3} = {r},
  cell{4}{4} = {r},
  cell{4}{5} = {r},
  cell{4}{6} = {r},
  cell{4}{7} = {r},
  cell{4}{8} = {r},
  cell{4}{9} = {r},
  cell{4}{10} = {r},
  cell{4}{11} = {r},
  cell{5}{2} = {r},
  cell{5}{3} = {r},
  cell{5}{4} = {r},
  cell{5}{5} = {r},
  cell{5}{6} = {r},
  cell{5}{7} = {r},
  cell{5}{8} = {r},
  cell{5}{9} = {r},
  cell{5}{10} = {r},
  cell{5}{11} = {r},
  cell{6}{2} = {r},
  cell{6}{3} = {r},
  cell{6}{4} = {r},
  cell{6}{5} = {r},
  cell{6}{6} = {r},
  cell{6}{7} = {r},
  cell{6}{8} = {r},
  cell{6}{9} = {r},
  cell{6}{10} = {r},
  cell{6}{11} = {r},
  cell{7}{2} = {r},
  cell{7}{3} = {r},
  cell{7}{4} = {r},
  cell{7}{5} = {r},
  cell{7}{6} = {r},
  cell{7}{7} = {r},
  cell{7}{8} = {r},
  cell{7}{9} = {r},
  cell{7}{10} = {r},
  cell{7}{11} = {r},
  cell{8}{2} = {r},
  cell{8}{3} = {r},
  cell{8}{4} = {r},
  cell{8}{5} = {r},
  cell{8}{6} = {r},
  cell{8}{7} = {r},
  cell{8}{8} = {r},
  cell{8}{9} = {r},
  cell{8}{10} = {r},
  cell{8}{11} = {r},
  cell{9}{2} = {r},
  cell{9}{3} = {r},
  cell{9}{4} = {r},
  cell{9}{5} = {r},
  cell{9}{6} = {r},
  cell{9}{7} = {r},
  cell{9}{8} = {r},
  cell{9}{9} = {r},
  cell{9}{10} = {r},
  cell{9}{11} = {r},
  cell{10}{2} = {r},
  cell{10}{3} = {r},
  cell{10}{4} = {r},
  cell{10}{5} = {r},
  cell{10}{6} = {r},
  cell{10}{7} = {r},
  cell{10}{8} = {r},
  cell{10}{9} = {r},
  cell{10}{10} = {r},
  cell{10}{11} = {r},
  vline{2-3} = {1-10}{},
  vline{4-11} = {2}{},
  vline{2-11} = {3-10}{},
  vline{6}   = {1-2}{},
  vline{9}   = {1-2}{},
  hline{1,3,11} = {-}{},
  hline{2} = {3-11}{},
}
Models        &  Accuracy & LAC      &       &            & APS      &       &            & RAPS     &       &            \\
              &           & Coverage & SSC   & $\bar{w}$ & Coverage & SSC   & $\bar{w}$ & Coverage & SSC   & $\bar{w}$ \\
ResNet18 &     0.825 &     0.941 & 0.818 &       1.561 &     0.961 & 0.920 &       2.250 &     0.960 & 0.920 &       2.233 \\
    ResNet34 &     0.839 &     0.939 & 0.934 &       1.484 &     0.958 & 0.936 &       2.085 &     0.958 & 0.933 &       2.138 \\
    ResNet50 &     0.838 &     0.940 & 0.857 &       1.493 &     0.959 & 0.800 &       2.027 &     0.961 & 0.857 &       2.148 \\
       VGG-16 &     0.835 &     0.946 & 0.875 &       1.530 &     0.970 & 0.952 &       2.274 &     0.970 & 0.951 &       2.318 \\
       VGG-19 &     0.823 &     0.943 & 0.935 &       1.561 &     0.964 & 0.875 &       2.222 &     0.966 & 0.923 &       2.316 \\
 DenseNet121 &     0.832 &     0.945 & 0.833 &       1.521 &     0.965 & 0.833 &       2.199 &     0.967 & 0.857 &       2.273 \\
 DenseNet161 &     0.836 &     0.943 & 0.938 &       1.496 &     0.962 & 0.932 &       2.208 &     0.964 & 0.929 &       2.283 \\
MobileNet-v2 &     0.827 &     0.944 & 0.940 &       1.563 &     0.965 & 0.952 &       2.237 &     0.966 & 0.952 &       2.311
\end{tblr}
\end{table*}

\begin{table*}[!h]
\centering
\caption{Models' and conformal predictors' performance on the ImageNet-ReaL dataset.}
\label{table:coverage_performance_imagenet-real}
\begin{tblr}{
  row{2} = {r},
  cell{1}{1} = {r=2}{},
  cell{1}{2} = {r=2}{},
  cell{1}{3} = {c=3}{c},
  cell{1}{6} = {c=3}{c},
  cell{1}{9} = {c=3}{c},
  cell{3}{2} = {r},
  cell{3}{3} = {r},
  cell{3}{4} = {r},
  cell{3}{5} = {r},
  cell{3}{6} = {r},
  cell{3}{7} = {r},
  cell{3}{8} = {r},
  cell{3}{9} = {r},
  cell{3}{10} = {r},
  cell{3}{11} = {r},
  cell{4}{2} = {r},
  cell{4}{3} = {r},
  cell{4}{4} = {r},
  cell{4}{5} = {r},
  cell{4}{6} = {r},
  cell{4}{7} = {r},
  cell{4}{8} = {r},
  cell{4}{9} = {r},
  cell{4}{10} = {r},
  cell{4}{11} = {r},
  cell{5}{2} = {r},
  cell{5}{3} = {r},
  cell{5}{4} = {r},
  cell{5}{5} = {r},
  cell{5}{6} = {r},
  cell{5}{7} = {r},
  cell{5}{8} = {r},
  cell{5}{9} = {r},
  cell{5}{10} = {r},
  cell{5}{11} = {r},
  cell{6}{2} = {r},
  cell{6}{3} = {r},
  cell{6}{4} = {r},
  cell{6}{5} = {r},
  cell{6}{6} = {r},
  cell{6}{7} = {r},
  cell{6}{8} = {r},
  cell{6}{9} = {r},
  cell{6}{10} = {r},
  cell{6}{11} = {r},
  cell{7}{2} = {r},
  cell{7}{3} = {r},
  cell{7}{4} = {r},
  cell{7}{5} = {r},
  cell{7}{6} = {r},
  cell{7}{7} = {r},
  cell{7}{8} = {r},
  cell{7}{9} = {r},
  cell{7}{10} = {r},
  cell{7}{11} = {r},
  cell{8}{2} = {r},
  cell{8}{3} = {r},
  cell{8}{4} = {r},
  cell{8}{5} = {r},
  cell{8}{6} = {r},
  cell{8}{7} = {r},
  cell{8}{8} = {r},
  cell{8}{9} = {r},
  cell{8}{10} = {r},
  cell{8}{11} = {r},
  cell{9}{2} = {r},
  cell{9}{3} = {r},
  cell{9}{4} = {r},
  cell{9}{5} = {r},
  cell{9}{6} = {r},
  cell{9}{7} = {r},
  cell{9}{8} = {r},
  cell{9}{9} = {r},
  cell{9}{10} = {r},
  cell{9}{11} = {r},
  cell{10}{2} = {r},
  cell{10}{3} = {r},
  cell{10}{4} = {r},
  cell{10}{5} = {r},
  cell{10}{6} = {r},
  cell{10}{7} = {r},
  cell{10}{8} = {r},
  cell{10}{9} = {r},
  cell{10}{10} = {r},
  cell{10}{11} = {r},
  vline{2-3} = {1-10}{},
  vline{4-11} = {2}{},
  vline{2-11} = {3-10}{},
  vline{6}   = {1-2}{},
  vline{9}   = {1-2}{},
  hline{1,3,11} = {-}{},
  hline{2} = {3-11}{},
}
Models        &  Accuracy & LAC      &       &            & APS      &       &            & RAPS     &       &            \\
              &           & Coverage & SSC   & $\bar{w}$ & Coverage & SSC   & $\bar{w}$ & Coverage & SSC   & $\bar{w}$ \\
ResNet18 &     0.725 &     0.902 & 0.500 &       2.994 &     0.939 &  0.000 &      12.680 &     0.939 &  0.000 &      12.712 \\
    ResNet34 &     0.759 &     0.899 & 0.681 &       2.157 &     0.940 &  0.000 &       8.894 &     0.940 &  0.500 &       8.943 \\
    ResNet50 &     0.785 &     0.903 & 0.000 &       1.760 &     0.946 &  0.000 &       7.768 &     0.946 &  0.000 &       7.892 \\
       VGG-16 &     0.745 &     0.902 & 0.000 &       2.461 &     0.943 &  0.000 &      10.690 &     0.943 &  0.000 &      10.587 \\
       VGG-19 &     0.751 &     0.903 & 0.000 &       2.355 &     0.946 &  0.000 &      10.764 &     0.946 &  0.000 &      10.692 \\
 DenseNet121 &     0.770 &     0.902 & 0.000 &       2.030 &     0.946 &  0.000 &       8.644 &     0.946 &  0.000 &       8.702 \\
 DenseNet161 &     0.796 &     0.902 & 0.000 &       1.633 &     0.942 &  0.000 &       6.053 &     0.942 &  0.000 &       6.034 \\
MobileNet-v2 &     0.744 &     0.902 & 0.688 &       2.496 &     0.940 &  0.000 &       9.539 &     0.941 &  0.000 &       9.730      
\end{tblr}
\end{table*}

\clearpage

\begin{table*}[h]
\centering
\caption{Similarity between conformal predictors and human annotations on the CIFAR-10H dataset}
\label{table:similarity_annotation_vs_cp_cifar10h}
\begin{tblr}{
  cells = {r},
  row{1} = {c},
  cell{1}{1} = {c=4}{},
  cell{1}{5} = {c=4}{},
  cell{1}{9} = {c=4}{},
  vline{2,6} = {1}{},
  vline{5,9} = {2-10}{},
  hline{1,11} = {-}{0.08em},
  hline{2} = {-}{},
  hline{3} = {-}{0.05em},
}
LAC       &        &          &                 & APS       &        &          &                 & RAPS      &        &          &                 \\
Precision & Recall & S. Acc & {Hamming\\loss ($\downarrow$)} & Precision & Recall & S. Acc & {Hamming\\loss ($\downarrow$)} & Precision & Recall & S. Acc & {Hamming\\loss ($\downarrow$)} \\
0.967     & 0.661  & 0.099    & 0.426           & 0.935     & 0.705  & 0.107    & 0.419           & 0.935     & 0.705  & 0.107    & 0.419           \\
0.966     & 0.667  & 0.099    & 0.429           & 0.936     & 0.707  & 0.109    & 0.415           & 0.941     & 0.702  & 0.107    & 0.418           \\
0.971     & 0.667  & 0.098    & 0.431           & 0.945     & 0.704  & 0.105    & 0.423           & 0.950     & 0.699  & 0.103    & 0.426           \\
0.972     & 0.667  & 0.098    & 0.429           & 0.943     & 0.707  & 0.105    & 0.422           & 0.948     & 0.703  & 0.103    & 0.424           \\
0.973     & 0.665  & 0.098    & 0.430           & 0.955     & 0.694  & 0.101    & 0.429           & 0.954     & 0.695  & 0.101    & 0.428           \\
0.974     & 0.656  & 0.099    & 0.428           & 0.951     & 0.697  & 0.105    & 0.425           & 0.952     & 0.696  & 0.104    & 0.425           \\
0.974     & 0.658  & 0.099    & 0.428           & 0.943     & 0.703  & 0.111    & 0.421           & 0.949     & 0.697  & 0.107    & 0.422           \\
0.975     & 0.664  & 0.098    & 0.431           & 0.939     & 0.709  & 0.107    & 0.421           & 0.939     & 0.709  & 0.107    & 0.422           
\end{tblr}
\end{table*}

\begin{table*}
\centering
\caption{Similarity between conformal predictors and human annotations on the MLRSNET dataset}
\label{table:similarity_annotation_vs_cp_mlrsnet}
\begin{tblr}{
  cells = {r},
  row{1} = {c},
  cell{1}{1} = {c=4}{},
  cell{1}{5} = {c=4}{},
  cell{1}{9} = {c=4}{},
  vline{2,6} = {1}{},
  vline{5,9} = {2-10}{},
  hline{1,11} = {-}{0.08em},
  hline{2} = {-}{},
  hline{3} = {-}{0.05em},
}
LAC       &        &          &                 & APS       &        &          &                 & RAPS      &        &          &                 \\
Precision & Recall & S. Acc & {Hamming\\loss ($\downarrow$)} & Precision & Recall & S. Acc & {Hamming\\loss ($\downarrow$)} & Precision & Recall & S. Acc & {Hamming\\loss ($\downarrow$)} \\
0.927     & 0.688  & 0.181    & 0.345        & 0.901     & 0.711  & 0.198    & 0.228        & 0.905     & 0.710  & 0.196    & 0.230        \\
0.931     & 0.665  & 0.173    & 0.351        & 0.889     & 0.720  & 0.192    & 0.219        & 0.884     & 0.722  & 0.197    & 0.218        \\
0.936     & 0.619  & 0.172    & 0.366        & 0.921     & 0.693  & 0.190    & 0.245        & 0.918     & 0.697  & 0.188    & 0.242        \\
0.941     & 0.655  & 0.166    & 0.352        & 0.900     & 0.728  & 0.180    & 0.209        & 0.907     & 0.709  & 0.184    & 0.206        \\
0.930     & 0.677  & 0.174    & 0.370        & 0.922     & 0.709  & 0.195    & 0.223        & 0.919     & 0.715  & 0.192    & 0.221        \\
0.931     & 0.682  & 0.160    & 0.351        & 0.899     & 0.698  & 0.201    & 0.208        & 0.892     & 0.697  & 0.199    & 0.207        \\
0.932     & 0.690  & 0.151    & 0.361        & 0.891     & 0.755  & 0.175    & 0.211        & 0.899     & 0.744  & 0.187    & 0.213        \\
0.933     & 0.674  & 0.163    & 0.373        & 0.881     & 0.699  & 0.202    & 0.201        & 0.895     & 0.723  & 0.201    & 0.198       
\end{tblr}
\end{table*}

\begin{table*}
\centering
\caption{Similarity between conformal predictors and human annotations on the FER+ dataset}
\label{table:similarity_annotation_vs_cp_ferplus}
\begin{tblr}{
  cells = {r},
  row{1} = {c},
  cell{1}{1} = {c=4}{},
  cell{1}{5} = {c=4}{},
  cell{1}{9} = {c=4}{},
  vline{2,6} = {1}{},
  vline{5,9} = {2-10}{},
  hline{1,11} = {-}{0.08em},
  hline{2} = {-}{},
  hline{3} = {-}{0.05em},
}
LAC       &        &          &                 & APS       &        &          &                 & RAPS      &        &          &                 \\
Precision & Recall & S. Acc & {Hamming\\loss ($\downarrow$)} & Precision & Recall & S. Acc & {Hamming\\loss ($\downarrow$)} & Precision & Recall & S. Acc & {Hamming\\loss ($\downarrow$)} \\
0.918     & 0.710  & 0.146    & 0.364        & 0.753     & 0.775  & 0.196    & 0.234        & 0.785     & 0.766  & 0.185    & 0.267        \\
0.934     & 0.698  & 0.144    & 0.365        & 0.756     & 0.762  & 0.186    & 0.214        & 0.754     & 0.763  & 0.187    & 0.213        \\
0.933     & 0.707  & 0.141    & 0.376        & 0.758     & 0.776  & 0.182    & 0.239        & 0.760     & 0.776  & 0.181    & 0.241        \\
0.934     & 0.717  & 0.137    & 0.384        & 0.756     & 0.795  & 0.186    & 0.249        & 0.761     & 0.794  & 0.185    & 0.253        \\
0.928     & 0.713  & 0.144    & 0.374        & 0.760     & 0.782  & 0.186    & 0.238        & 0.740     & 0.788  & 0.192    & 0.224        \\
0.928     & 0.712  & 0.139    & 0.377        & 0.740     & 0.784  & 0.190    & 0.217        & 0.739     & 0.784  & 0.191    & 0.217        \\
0.936     & 0.713  & 0.136    & 0.386        & 0.727     & 0.785  & 0.190    & 0.209        & 0.723     & 0.786  & 0.192    & 0.205        \\
0.918     & 0.716  & 0.143    & 0.375        & 0.758     & 0.782  & 0.194    & 0.242        & 0.765     & 0.781  & 0.191    & 0.252        
\end{tblr}
\end{table*}

\begin{table*}[!t]
\centering
\caption{Similarity between conformal predictors and human annotations on the ImageNet-ReaL dataset}
\label{table:similarity_annotation_vs_cp_imagenetreal}
\begin{tblr}{
  cells = {r},
  row{1} = {c},
  cell{1}{1} = {c=4}{},
  cell{1}{5} = {c=4}{},
  cell{1}{9} = {c=4}{},
  vline{2,6} = {1}{},
  vline{5,9} = {2-10}{},
  hline{1,11} = {-}{0.08em},
  hline{2} = {-}{},
  hline{3} = {-}{0.05em},
}
LAC       &        &          &                 & APS       &        &          &                 & RAPS      &        &          &                 \\
Precision & Recall & S. Acc & {Hamming\\loss ($\downarrow$)} & Precision & Recall & S. Acc & {Hamming\\loss ($\downarrow$)} & Precision & Recall & S. Acc & {Hamming\\loss ($\downarrow)$} \\
0.571     & 0.892  & 0.002    & 0.370        & 0.568     & 0.934  & 0.012    & 0.417        & 0.567     & 0.934  & 0.012    & 0.416        \\
0.670     & 0.885  & 0.001    & 0.481        & 0.631     & 0.930  & 0.008    & 0.480        & 0.629     & 0.931  & 0.008    & 0.478        \\
0.736     & 0.882  & 0.001    & 0.559        & 0.678     & 0.931  & 0.007    & 0.529        & 0.679     & 0.931  & 0.007    & 0.529        \\
0.628     & 0.887  & 0.002    & 0.434        & 0.604     & 0.933  & 0.010    & 0.453        & 0.603     & 0.934  & 0.010    & 0.452        \\
0.643     & 0.886  & 0.002    & 0.452        & 0.611     & 0.934  & 0.010    & 0.462        & 0.612     & 0.934  & 0.010    & 0.462        \\
0.684     & 0.884  & 0.001    & 0.494        & 0.632     & 0.934  & 0.008    & 0.479        & 0.632     & 0.934  & 0.008    & 0.479        \\
0.763     & 0.876  & 0.001    & 0.593        & 0.705     & 0.921  & 0.005    & 0.552        & 0.702     & 0.923  & 0.005    & 0.548        \\
0.625     & 0.887  & 0.002    & 0.430        & 0.604     & 0.930  & 0.008    & 0.452        & 0.599     & 0.932  & 0.009    & 0.447        
\end{tblr}
\end{table*}

\begin{table*}
\centering
\caption{Spearman's rank correlation coefficient, $r_s$, $p < .001$, between conformal prediction set sizes and human annotation entropy.}
\label{table:spearman_cp_sets_and_human_annotation_entropy}
\begin{tblr}{
  row{2} = {r},
  cell{1}{1} = {r=2}{},
  cell{1}{2} = {c=3}{c},
  cell{1}{5} = {c=3}{c},
  cell{3}{2} = {r},
  cell{3}{3} = {r},
  cell{3}{4} = {r},
  cell{3}{5} = {r},
  cell{3}{6} = {r},
  cell{3}{7} = {r},
  cell{4}{2} = {r},
  cell{4}{3} = {r},
  cell{4}{4} = {r},
  cell{4}{5} = {r},
  cell{4}{6} = {r},
  cell{4}{7} = {r},
  cell{5}{2} = {r},
  cell{5}{3} = {r},
  cell{5}{4} = {r},
  cell{5}{5} = {r},
  cell{5}{6} = {r},
  cell{5}{7} = {r},
  cell{6}{2} = {r},
  cell{6}{3} = {r},
  cell{6}{4} = {r},
  cell{6}{5} = {r},
  cell{6}{6} = {r},
  cell{6}{7} = {r},
  cell{7}{2} = {r},
  cell{7}{3} = {r},
  cell{7}{4} = {r},
  cell{7}{5} = {r},
  cell{7}{6} = {r},
  cell{7}{7} = {r},
  cell{8}{2} = {r},
  cell{8}{3} = {r},
  cell{8}{4} = {r},
  cell{8}{5} = {r},
  cell{8}{6} = {r},
  cell{8}{7} = {r},
  cell{9}{2} = {r},
  cell{9}{3} = {r},
  cell{9}{4} = {r},
  cell{9}{5} = {r},
  cell{9}{6} = {r},
  cell{9}{7} = {r},
  cell{10}{2} = {r},
  cell{10}{3} = {r},
  cell{10}{4} = {r},
  cell{10}{5} = {r},
  cell{10}{6} = {r},
  cell{10}{7} = {r},
  vline{2} = {1-2}{},
  vline{5} = {2-3}{},
  vline{2,5} = {3-10}{},
  hline{1,11} = {-}{0.08em},
  hline{2} = {2-7}{},
  hline{3} = {1-2}{},
  hline{3} = {3-7}{0.03em},
}
 Models      & CIFAR-10H &       &       & FER+  &       &       \\
             & LAC       & APS   & RAPS  & LAC   & APS   & RAPS  \\
ResNet18     & 0.140     & 0.255 & 0.255 & 0.404 & 0.354 & 0.365 \\
ResNet34     & 0.142     & 0.233 & 0.228 & 0.413 & 0.272 & 0.272 \\
ResNet50     & 0.160     & 0.246 & 0.240 & 0.413 & 0.309 & 0.310 \\
VGG-16       & 0.153     & 0.254 & 0.246 & 0.457 & 0.297 & 0.301 \\
VGG-19       & 0.162     & 0.240 & 0.241 & 0.429 & 0.296 & 0.276 \\
DenseNet121  & 0.025     & 0.230 & 0.228 & 0.437 & 0.191 & 0.191 \\
DenseNet161  & 0.058     & 0.228 & 0.220 & 0.454 & 0.277 & 0.272 \\
MobileNet-v2 & 0.124     & 0.269 & 0.268 & 0.427 & 0.337 & 0.343 
\end{tblr}
\end{table*}

\begin{table*}[!h]
\centering
\caption{Expected Calibration Error (ECE) with M = 15 bins.}
\label{table:ece}
\begin{tblr}{
  column{even} = {r},
  column{3} = {r},
  column{5} = {r},
  vline{2-5} = {-}{},
  hline{1-2,10} = {-}{},
}
Models       & CIFAR-10H & MLRSNet & FER+  & ImageNet-ReaL \\
ResNet18     & 0.021         & 0.013   & 0.032 & 0.021             \\
ResNet34     & 0.026         & 0.023   & 0.035 & 0.027             \\
ResNet50     & 0.023         & 0.030   & 0.026 & 0.032             \\
VGG-16       & 0.016         & 0.100   & 0.037 & 0.020             \\
VGG-19       & 0.022         & 0.119   & 0.053 & 0.020             \\
DenseNet121  & 0.020         & 0.018   & 0.034 & 0.020             \\
DenseNet161  & 0.021         & 0.019   & 0.038 & 0.047             \\
MobileNet-v2 & 0.025         & 0.012   & 0.027 & 0.020             
\end{tblr}
\end{table*}

\end{document}